\documentclass{article}

\PassOptionsToPackage{numbers, compress}{natbib}

 \usepackage[main, final]{neurips_2026}

\usepackage[utf8]{inputenc} 
\usepackage[T1]{fontenc}    
\usepackage{hyperref}       
\usepackage{url}            
\usepackage{booktabs}       
\usepackage{amsfonts}       
\usepackage{nicefrac}       
\usepackage{microtype}      
\usepackage{xcolor}         

\usepackage{float} 
\usepackage{wrapfig}
\usepackage{tikz}
\usetikzlibrary{fit, arrows, positioning, bending, shapes, shapes.geometric, arrows.meta, calc}
\usepackage[inline]{enumitem}

\usepackage[most]{tcolorbox} 
\newtcolorbox{definitionbox}{
  colback=gray!10,    
  colframe=gray!80,   
  boxrule=0.5pt,      
  arc=2mm,            
  left=6pt, right=6pt, top=6pt, bottom=6pt, 
}

\newtheorem{example}{Example}

\definecolor{Purple}{HTML}{800080}
\definecolor{DarkGreen}{RGB}{  0,115,  0}   
\definecolor{DarkRed}  {RGB}{140,  0,  0}   
\definecolor{mygreen}{RGB}{34,139,34}

\newcommand{\our}{\textcolor{black}{DeepObjectLog}}

\newcommand{\unavailable}{-}

\title{ Neurosymbolic Object-Centric Learning \\with Distant Supervision}

\author{%
  Stefano Colamonaco \\
  KU Leuven\\
  \texttt{stefano.colamonaco@kuleuven.be} \\
  \And
  David Debot \\
  KU Leuven \\
  \texttt{david.debot@kuleuven.be} \\
  \AND
  Giuseppe Marra\\
  KU Leuven \\
  \texttt{giuseppe.marra@kuleuven.be} \\
}

\begin{document}

\maketitle

\begin{abstract}

Neurosymbolic learning can use symbolic rules to provide supervision for latent concepts from weak labels, but it commonly assumes that the entities referenced by these rules are already specified. Object-centric models decompose images into slot-like representations; however, such slots are not necessarily aligned with the predicates required for symbolic reasoning. We investigate object-centric neurosymbolic learning under distant supervision, where the object-level arguments of a logic program are learned directly from images using only global task labels. We introduce \our{}, a probabilistic neurosymbolic model that integrates a slot-based perceptual encoder with a probabilistic logic layer. The encoder predicts objectness and class probabilities for candidate object representations, while the logic layer marginalizes over latent objectness and class assignments to compute the likelihood of the observed label. This formulation provides a differentiable task-level learning signal for object-centric perception without requiring per-object labels, masks, bounding boxes, or heuristic set matching. Evaluations across diverse visual reasoning tasks demonstrate that \our{} achieves superior out-of-distribution generalization to compositional, object-count, and rule shifts compared to neural object-centric and standard neurosymbolic baselines.
\end{abstract}

\addtocontents{toc}{\protect\setcounter{tocdepth}{-1}}

\section{Introduction}

Neurosymbolic AI aims to combine neural representation learning with explicit reasoning over entities, relations, and compositional rules \citep{manhaeve2018deepproblog,skryagin2022neural,marra2024statistical}. A central appeal of this paradigm is that symbolic knowledge can provide supervision for latent concepts even when direct labels are unavailable. For example, a model trained only on the sum of two digit images can learn to classify the individual digits, because the addition rule constrains which latent digit assignments are compatible with the observed sum \citep{manhaeve2018deepproblog}. However, this ability usually relies on an important precondition: the entities over which reasoning should occur are already given. In the digit-addition example, the model is typically provided with two separate digit images. It must learn what each digit is, but not where the digits are, how many there are, or which parts of a larger image correspond to distinct objects.

This precondition limits the applicability of neurosymbolic learning to perceptual inputs whose object structure is not known in advance. In many visual domains, an image may contain an unknown number of entities, with no labels, masks or bounding boxes. Before a symbolic rule can constrain object properties, the model must infer which candidate parts of the input should be treated as task-relevant objects. Thus, the bottleneck is not only learning symbolic concepts from weak supervision, but also learning the object-centric structure that makes symbolic reasoning possible.

Object-centric learning addresses this complementary problem by decomposing visual scenes into slot-like representations, each intended to capture a candidate object or scene component \citep{burgess2019monet,greff2019multi,locatello2020object,de2024object}. These representations provide a natural interface for neurosymbolic reasoning, since slots can in principle serve as the entities over which rules operate. Yet slots are not automatically aligned with the predicates needed by a symbolic task. A slot may capture a visual region without being identifiable as a task-relevant object, and a set of slots may support a neural classifier without producing the explicit object-level structure required by the symbolic component. When object-level supervision is unavailable, it remains unclear how to align object-centric perception with symbolic reasoning.

We study this interface. Given image-level labels and a program describing how latent object-level predicates determine those labels, we ask whether symbolic task knowledge can guide the learning of task-relevant object-centric structure, as specified in the box below.

\begin{definitionbox}
\textbf{Problem. [\textit{Object-centric neurosymbolic learning from distant supervision}]}
\newline

\noindent \textbf{\textit{Setup}:} We are given a dataset of image--label pairs \((x,y)\), where each image may contain multiple objects but no annotation is provided for object identity, location, mask, or class. We are also given a probabilistic logic program \(L\), a set of rules specifying how latent object-level predicates, e.g. objectness and class, determine the global label \(y\). Notably, the label alone does not reveal how many objects are present, where they are, or which classes they belong to and the given symbolic rules only constrain which decompositions could explain the observation.

\noindent \textbf{\textit{Goal}:} Learn a neurosymbolic model in which perception and reasoning inform each other: the perceptual module proposes candidate objects and their predicates, while the symbolic layer evaluates whether those predicates can explain the observed task label. The learned model should support relational generalization under distribution shifts, including new object compositions, different object counts within the slot capacity of the model, and new supplied symbolic queries or rules over the learned object-level vocabulary.
\end{definitionbox}

We propose \our{}, a neurosymbolic architecture that integrates object-centric perception with probabilistic logical reasoning. The perceptual component extracts a  set of candidate object representations from an image. For each candidate, the model predicts an objectness probability and a distribution over symbolic classes. These predictions are passed as probabilistic facts to a ProbLog program \citep{de2007problog}, which computes the marginal probability of the observed task label by tractably summing over possible objectness and class assignments. This marginal likelihood provides a differentiable signal from the global task label back to the object-centric encoder, allowing logical constraints to influence the selection, classification, and grounding of candidate objects.

We evaluate \our{} on MultiMNIST-Addition, PokerRules, CLEVR-Addition, and a COCOLogic-based setting \citep{steinmann2025object}. These controlled benchmarks test whether a model can use task-level symbolic structure to learn object-centric representations that support reasoning under distribution shift. Across these settings, \our{} achieves stronger out-of-distribution task performance than neural object-centric baselines and than a neurosymbolic baseline using the same logic but lacking object-centric perception.

The contributions of this work are threefold. First, we formulate object-centric neurosymbolic learning from distant supervision, where objects are not provided as structured inputs but must be inferred from perception. Second, we introduce a probabilistic interface between slots and logic, in which objectness and class variables are marginalized by probabilistic logical inference, providing a differentiable task-level signal without object annotations, discrete sampling, or set-matching losses. Third, we instantiate this formulation in \our{} and show that logical feedback improves the task-relevant use of object-centric representations and supports stronger relational generalization, while also analyzing limitations such as inference cost, slot-capacity sensitivity, and cases where task-level labels do not uniquely identify human-named concepts.

\begin{figure*}
    \centering
    \includegraphics[width=1.0\linewidth]{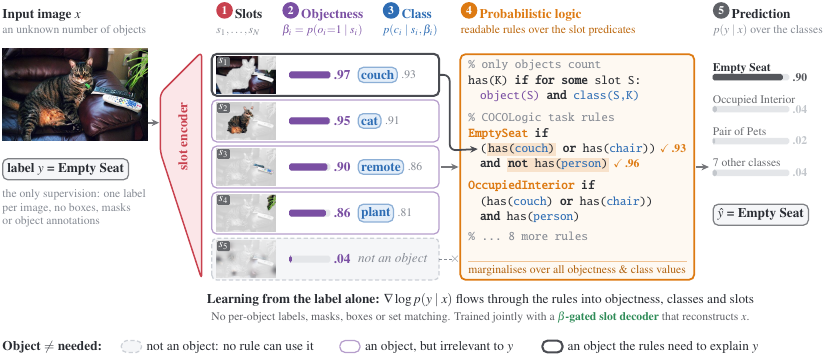}
    \caption{ 
    Overview of the proposed model, \our, on a COCOLogic example. A slot encoder extracts N slots, and each slot gets an objectness probability and a class distribution. A logic program marginalizes over all objectness and class assignments to compute the probability of each task label, so the label alone supervises objectness, classes and slots, with no per-object labels, masks, boxes or set matching. Objectness also gates the slot decoder, so a slot that is not an object (s5) is used by neither the logic nor the reconstruction.
    }
    \label{fig:model}
\end{figure*}

A graphical overview of the proposed model is provided in Figure~\ref{fig:model}, while a detailed explanation of its components is given in Section~\ref{sec:method}.


\section{Related Work}
The problem addressed in this study lies at the intersection of two research areas: object-centric representation learning and neurosymbolic reasoning. Both are essential, but neither is sufficient on its own to meet the demands of a system capable of learning relational abstractions directly from raw input  when supervision is provided only at the task level.

\textbf{Object-centric learning.}
Object-centric models aim to decompose visual scenes into discrete object representations \citep{Greff2020}. Early approaches \citep{burgess2019monet, greff2019multi, engelcke2019genesis} use iterative variational inference to discover object structure in an unsupervised manner, but often suffer from scalability limitations. Slot Attention \citep{locatello2020object} replaces these procedures with a parallel attention-based binding mechanism, enabling permutation-invariant inference over latent slots. This model has since been extended across several domains \citep{kori2023grounded, elsayed2022savi++, seitzer2022bridging, singh2022simple}, with slot-based representations being applied to generative modeling and dynamic scenes. 
Some works attempt to improve the selection of useful slots by adaptively selecting which representations carry meaningful content \citep{fan2024adaptive, engelcke2021genesisv2}, but these still rely on heuristics or sparsity priors that are disconnected from task-level semantics. A recent study \citep{kori2023grounded} shows that slot-based models can be trained end-to-end with supervision on a downstream task, but it assumes a fixed number of objects and employs a neural classifier without symbolic reasoning, requiring few-shot learning to achieve task generalization. 
While certain architectures perform downstream tasks like VQA without explicit labels \citep{wu2022slotformer}, they often lack discrete symbolic grounding. Without task-level logical supervision, current models learn representations independent of the task's symbolic structure, limiting their interpretability and out-of-distribution generalization \citep{wiedemer2023provable}.

\textbf{Neurosymbolic systems. } A complementary limitation appears in most neurosymbolic (NeSy) systems: while they reason over structured entities, they assume that this structure is already available. Thus, they cannot learn to infer object-centric representations from raw data. While NeSy approaches integrate neural perception with symbolic logic for probabilistic reasoning \citep{marra2024statistical, manhaeve2018deepproblog}, they typically assume inputs consist of pre-segmented objects (e.g. individual images or attribute vectors), bypassing the challenge of structure discovery \citep{yi2018neural, mao2019neuro}. Existing extensions incorporate limited spatial reasoning or component-level architectures but usually operate with a fixed number of entities \citep{misino2022vael, de2023neural}. Consequently, these systems lack object-agnostic parameterization.

\citet{stammer2021right} and \citet{shindo2023alpha} employ a pretrained Slot Attention module for object-centric representations, subsequently using them for rule extraction. \citet{skryagin2022neural} explore the integration of logical constraints into the training of object-centric modules, demonstrating that logical supervision can improve object decomposition by introducing assumptions of independence across attributes. Finally, some works in abductive learning have attempted to learn object detection from unstructured data, but they typically rely on knowing in advance how many objects are present in each image, which limits their ability to fully handle unstructured inputs \citep{gao2024knowledge, cai2021abductive}. While these approaches demonstrate the potential of combining perception with logic, they still rely on object-level supervision, such as object labels or question-answer pairs, to align neural representations with symbolic targets.

\section{Method} \label{sec:method}


To address these limitations, we introduce \our{}, a neurosymbolic architecture that learns object-centric representations from weak supervision and reasons over their latent properties.

\begin{wrapfigure}{r}{0.35\linewidth}
    \centering
        \begin{tikzpicture}[->, >=stealth, node distance=0.8cm]
            \begin{scope}[scale=0.70, transform shape]
                \node (Z) [draw, circle] {$z$};
                \node (S) [draw, circle, right=of Z)] {$s_i$};
                \node (B) [draw, circle, right=of S)] {$\beta_i$};
                \node (C) [draw, circle, right=of B, yshift=1.0cm)] {$c_i$};
                \node (O) [draw, circle, below=of C, yshift=-0.4cm )] {$o_i$};
                \node (y) [draw, circle,minimum width=0.7cm, minimum height=0.7cm, right=of C, yshift=-1cm)] {$y$};
                \node (x) [draw, circle, below=of B, yshift=-0.8cm )] {$x$};
    
                \node[draw, dashed, thick, fit=(S)(O)(C), inner sep=0.35cm] (plate1) {};
                \node[anchor=north west] at (plate1.north west) {$i = 1..N$};
    
                \draw[draw=red, thick] (Z) -- (S);
                \draw[draw=Purple, thick] (S) -- (B);
                \draw[draw=Purple, thick] (B) -- (O);
                \draw[draw=blue, thick] (B) -- (C);
                \draw[draw=orange, thick] (O) -- (y);
                \draw[draw=orange, thick] (C) -- (y);

                \draw[draw=DarkGreen, thick] (B) -- (x);
                \draw[draw=DarkGreen, thick] (S) -- (x);
    
                
                \coordinate (exitpoint1) at ($(S)+(0,1)$);
                \draw[draw=blue, thick] (S) -- (exitpoint1) -- (C);

                
            \end{scope}
        \end{tikzpicture}
        \caption{Probabilistic graphical model of \our}
        \label{fig:pgm}
\end{wrapfigure}
From a high level, the architecture is composed of the following components:
(i) \textcolor{red}{\textbf{Objects encoder}}: starting from a global representation $z$, the model extracts up to $N$ latent representations $s_i$, each intended to describe a potential object in the scene; 
(ii) \textbf{Object \textcolor{Purple}{detector} and \textcolor{blue}{classifier}}: for each extracted object encoding, the model infers a probability $\beta_i$ that the representation corresponds to a meaningful object, from which an \textit{objectness} flag $o_i$ and a class $c_i$ are extracted;
(iii) \textcolor{orange}{\textbf{Probabilistic logic reasoning}}: the system reasons logically over detected objects and their classes to derive a downstream task. This enables the system to interpret objects, understand their relationships, and constrain object configurations in ways that support task-specific inference.
(iv) \textcolor{DarkGreen}{\textbf{Image decoder}}: finally, each representation is decoded into image space.

\subsection{Probabilistic graphical model}
Let $z \in \mathbb{R}^{d_z}$ be a latent representation of the image. For each potentially present object $i$, we denote its associated slot as $s_i \in \mathbb{R}^{d_s}$, and the probability that it represents an actual object as $\beta_i \in [0,1]$. The corresponding objectness flag is $o_i \in \{0,1\}$ and the class label corresponding to the object is $c_i \in C \subset \mathbb{N}$. Finally, let $y \in Y \subset \mathbb{N}$ be the downstream class label computed by a symbolic logic theory $L$, and  $x \in [0,1]^{ch \times h \times w}$ the normalized image.
The generative model in Figure \ref{fig:pgm} factorizes as:
\begin{align}
p\big(y,x,\mathbf{s},\boldsymbol{\beta},\mathbf{o},\mathbf{c},z\big)
  \;=\;
  p(z)\;
    p_L \bigl(y|\mathbf{o},\mathbf{c}\bigr)\;
  p \bigl(x| \mathbf{s},\boldsymbol{\beta}\bigr) 
  \cdot \Bigl[\prod_{i=1}^{N} p(s_i\mid z)p(\beta_i | s_i) \,p(o_i| \beta_i)\,p(c_i| \beta_i,s_i)\Bigr]\;
\label{eq:joint}
\end{align}
where 
\(\mathbf{s}=(s_1,\dots,s_N),\;
  \boldsymbol{\beta}=(\beta_1,\dots,\beta_N),\;
  \mathbf{o}=(o_1,\dots,o_N),\;
  \mathbf{c}=(c_1,\dots,c_N)\). 

This joint distribution consists of the following components:
\begin{itemize}[leftmargin=5.5mm]
    \item $p(z)$ a prior distribution of the latent representation of the scene; 
    \item $p(s_i \mid z)$: the \textit{slot extractor}, which produces the $i$-th latent slot representation $s_i$ out of $N$ maximum possible objects based on the latent representation $z$. A slot $s_i$ represents a \textit{proposal} for a potential object.
    \item $p(\beta_i \mid s_i)$ models the uncertainty of $s_i$ being a meaningful object (and not, e.g., part of the background); it is represented explicitly since object-centric models can condition their processing on such uncertainty \citep{fan2024adaptive};
    \item $p(\text{o}_i \mid \beta_i)$ models the binary flag corresponding to the uncertainty $\beta_i$, 
    \item $p(c_i \mid \beta_i, s_i)$: the symbolic classification head, which assigns a discrete class label $c_i$ to the representation $s_i$, conditioned on its uncertain objectness $\beta_i$; 
    \item $p_L\left(y \mid \mathbf{o}, \mathbf{c} \right)$: the symbolic reasoning component, which predicts the final task-specific output $y$ by reasoning on a logic theory $L$, the inferred objects $o_i$ and their classes $c_i$;
    \item $p(x \mid \mathbf{s}, \boldsymbol{\beta})$: the image generator, a probability distribution over images given the set of slots and their predicted objectness.
\end{itemize}

This formulation makes object structure latent and uncertain, while coupling symbolic prediction and perceptual reconstruction through shared object-centric representations.
Notice that, although $N$ fixes the maximum number of slots, all slot-wise modules share parameters and are independent of the actual number of objects, which is inferred through the objectness variables.

In the sections below, we describe the main components of the system in more detail.

\subsection{Object independent encoder in the discriminative setting}
In this paper, we are interested in a discriminative setting, where we focus on modeling the probability distribution of the object classes $c_i$ by observing the task  label $y$ (see Section \ref{sec:learning}), and using the image $x$ to amortize inference.  Following  \citet{kori2024identifiable}, we model the slot encoder distributions $p(s_i | \cdot)$ as deterministic delta distributions. 
Such an encoder should avoid relying on supervision or assumptions about object count or location, ensuring flexibility across inputs and generalization to novel object numbers and positions.
Many possible existing models satisfy such requirements \citep{burgess2019monet,engelcke2019genesis} but we leave their comparison for future work. In our implementation, we exploit Slot Attention \citep{locatello2020object,zhang2023unlocking}, a differentiable mechanism for object-centric representation learning. In particular, $p(s_i | z) = SlotAttention_i(z; \theta_s, N)$
where $\theta_s$ is the real set of parameters of the slot attention model, which are shared across slots. The number $N$ is therefore only a maximum-slot hyperparameter, while downstream objectness variables determine which slots correspond to actual objects.

\subsection{Object detector and classifier}
\label{sec:detector_classifier}
The object detection is modelled over two variables: the uncertainty scores $\beta_i \in [0,1]$ and the objectness flags $o_i \in \{0,1\}$. The meaning is that $o_i$ is 1 (i.e. it represents an object) with probability $\beta_i$.  We model directly the uncertainty $\beta_i$ over $o_i$ as many components in current state-of-the-art object-centric models exploit soft masking schemes using such uncertainty. However, we do not model any probability distribution over $\beta_i$; therefore $p(\beta_i | s_i) = \delta(\beta_i - \beta^*_i)$ is a delta distribution centered around the output value of a neural network $\beta^*_i = f_\beta(s_i | \theta_\beta)$ with $\theta_\beta$ its set of weights. 
We model $p(o_i | \beta_i)$ as a Bernoulli distribution, parameterized by $\beta_i$, i.e. $p(o_i | \beta_i) = \beta_i$ . 
The classifier $p(c_i | s_i, \beta_i)$ can be parameterized with either a Bernoulli or with a Categorical distribution, depending on the particular object classification task. In both cases, the distribution is parameterized by 
a neural network $f_c(\beta_is_i| \theta_c)$ with weights $\theta_c$.
Following \citet{fan2024adaptive}, the slot representation $s_i$ in the input to the network $f_c$ is also multiplied elementwise by the uncertainty $\beta_i$. 
gating uncertain slots and reducing the influence of background or spurious proposals.

\subsection{Probabilistic Logic Reasoning over the tasks}
The final symbolic prediction $y$ is computed using probabilistic logical reasoning over the set of inferred object class predictions and objectness probabilities. This component is implemented using ProbLog \citep{de2007problog}, a probabilistic logic programming framework that extends classical logic programming with uncertainty modeling through probabilistic facts. While ProbLog is Turing-equivalent and allows for highly expressive probabilistic modeling, in this work we focus only on the aspects relevant to neurosymbolic object-centric learning and limit our explanation accordingly. For a full overview of the framework, we refer interested readers to \citet{de2007problog}.

\paragraph{Probabilistic Logic Programming.} A ProbLog program $L$ is defined over two sets of syntactic constructs: probabilistic facts and rules. A probabilistic fact with syntax \texttt{p::f} is an independent Bernoulli distribution $p(f)$ over a binary variable \texttt{f} with parameter $\texttt{p}=p(f)$. For example, \texttt{0.1::alarm} means that variable \texttt{alarm} has 0.1 probability of being True. A rule is of the form $\texttt{h :- }\texttt{b}_1, ..., \texttt{b}_m$, where \texttt{h} is the head or conclusion and each $\texttt{b}_i$ is a body element or premise. Rules act as definition rules: if all the premises $\texttt{b}_i$ are True, then the conclusion $\texttt{h}$ must be True. For example, in the classical Pearl's burglary example, the rule "\texttt{call :- alarm, hear}" encodes that if the \texttt{alarm} goes off and we \texttt{hear} it, then we \texttt{call} the police. 
Given the set of all probabilistic facts $\mathbf{f}$ and the set of all possible rules $R$, ProbLog allows to efficiently compute the marginal probability of every symbol $y$ in the program as:
\begin{equation}
\label{eq:problog_inference}
p(y) = \sum_\mathbf{f} p(y | \mathbf{f}; R) \prod_{f_i \in \mathbf{f}} p(f_i)
\end{equation}
where $p(y | \mathbf{f}; R)$ is a deterministic distribution stating whether $y$ can be obtained from $\mathbf{f}$ when applying the rules in $R$ (possibly chaining multiple of them).

\paragraph{ProbLog for object-centric learning.} We use ProbLog to model three components of our model:
\begin{itemize}[leftmargin=5.5mm]
	\item The object detectors $p(o_i | \beta_i)$ are modelled as probabilistic facts \texttt{object(i)}
	stating whether the i-th extracted slot   is an object;
	\item The object classifiers $p(c_i | s_i, \beta_i)$ are modelled as probabilistic facts \texttt{class(i)} (for a binary classifier) or \texttt{class(i,k)} for a categorical classifier with class $k$. The conditional nature on the objectness $o_i$ is encoded using rules. The rule \texttt{task:-object(1),class(1,3)}, for example, states that \texttt{task} is true only if the $1$st slot is an object and its class is $3$. 
	\item The conditional task distribution $p_L(y | \mathbf{c}, \mathbf{o})$  is the deterministic distribution obtained by all the logic rules proving the task $y$ given the objectnesses and the classes of all objects.
\end{itemize}

\begin{example}[Addition of at most two digits] 
Let us consider a setting where images contain zero to two digits that can only be $0$ or $1$. Consider the following two kinds of probabilistic facts. First,  $\mathtt{p_{o_{id}} :: object(ID)}$ stating that the probability that the $\mathtt{ID}$-th slot of the input image contains an actual digit is $\mathtt{p_{o_{id}}}$, as computed by $p(o_i | \beta_i)$. Second,   $\mathtt{p_{c_{id}} :: class(ID, C)}$ providing the probability that the $\mathtt{ID}$-th object is of class $C$, as computed by $p(c_i | s_i, \beta_i)$. 
We can encode the digit addition task in the following program:
\begin{verbatim}
digit(ID, Val) :- object(ID), class(ID, Val). 
digit(ID, 0) :- \+ object(ID). 
add(Z) :- digit(1, Y1), digit(2, Y2), Z is Y1 + Y2.
\end{verbatim}
We can then query, for example, what is the probability that a given image with multiple digits sums to 2, i.e. $y = \mathtt{add(2)}$, given the distributions of the current objectness of the slots and their corresponding classes. 
\end{example}

Note that any marginal task distribution $p(y | \mathbf{s}, \boldsymbol{\beta})$, can be computed by standard marginal inference (Eq. \ref{eq:problog_inference}) of the task query $y$:
\begin{equation}
\label{eq:task_marginal} p(y | \mathbf{s}, \boldsymbol{\beta}) = \sum_{\mathbf{c}, \mathbf{o}} p(y | \mathbf{c}, \mathbf{o}) \prod_i p(o_i | \beta_i) p(c_i | s_i, \beta_i)
\end{equation}
It is interesting to note that the use of probabilistic logic eliminates the need for set matching losses typically employed in object-centric systems (e.g., Hungarian loss), as object assignments are implicitly resolved through marginalization.

\paragraph{On the Role of Knowledge.}
The logic rules in our framework describe how object-level concepts should be composed to determine the global label, yet they do not provide any per-instance supervision. This is standard in neurosymbolic learning and reflects the core premise of the paradigm: when symbolic knowledge is available, it should be seamlessly usable by the model. If no knowledge is provided, the model trivially reduces to a purely neural architecture, as standard deep approaches are simply a special case within our framework. Conversely, when partial knowledge exist (e.g., from human experts, domain rules, knowledge bases, or even rules induced by LLMs \cite{shindo2024deisam}) the same interface allows it to be incorporated naturally and consistently.

\subsection{Masked Decoder}
For the image decoder $p(x | \mathbf{s}, \boldsymbol{\beta})$, we follow recent literature in Slot-attention architectures. In particular, we model the image as a mixture of Multivariate Gaussian distributions. Each component is a multivariate Gaussian $\mathcal{N}(\mu, I)$ with unit covariance matrix $I$ and a mean vector $\mu$ parameterized with a neural network $f_x(s_i \beta_i ; \theta_x)$ with $\theta_x$ its parameters. The mixing weights $w$ of the components are parameterized by a neural network $f_w(s_i \beta_i ; \theta_w)$. Notice, as we did for the objectness variable $o_i$ and as often done in the slot-attention literature, we decode each slot $s_i$ after a soft masking using $\beta_i$.

\section{Learning}
\label{sec:learning}

In the discriminative setting considered in this work, learning is
formulated as joint optimization over model parameters and latent explanations.
Each training image--label pair \((x_k,y_k)\) is assigned a single latent
explanation \(z_k\), and the model is trained to maximize the joint probability
of the observed data and their explanations:
\begin{equation}
\label{eq:ideal_map_learning}
\max_{\theta_p,\{z_k\}_{k=1}^K}
\sum_{k=1}^K
\log p_{\theta_p}(x_k,y_k,z_k).
\end{equation}
This is a MAP-style objective, in which the model searches for one high-scoring explanation per training
example. Here \(z_k\) is the latent explanation for each training example \((x_k, y_k)\)  and \(\theta_p=\{\theta_s,\theta_\beta,\theta_c,\theta_x,\theta_w\}\) denotes
the parameters of the probabilistic model. 

Starting from the full joint model
\(
p_{\theta_p}(y,x,\mathbf{s},\boldsymbol{\beta},\mathbf{o},\mathbf{c},z),
\)
the score \(p_{\theta_p}(x,y,z)\) is obtained by marginalizing the internal
variables that are not directly optimized.
Since \(\mathbf{s}\) and \(\boldsymbol{\beta}\) are deterministic in our model,
their distributions collapse to point values
\(\mathbf{s}^\star(z)\) and \(\boldsymbol{\beta}^\star(z)\). Using the sifting
property of delta distributions and the task marginalization of ProbLog
(Equation~\ref{eq:task_marginal}), we obtain
\begin{equation}
\label{eq:collapsed_joint_score}
p_{\theta_p}(x,y,z)
=
p(z)\,
p_{\theta_p}\big(
x\mid \mathbf{s}^\star(z),\boldsymbol{\beta}^\star(z)
\big)\,
p_{\theta_p}\big(
y\mid \mathbf{s}^\star(z),\boldsymbol{\beta}^\star(z)
\big).
\end{equation}

The ideal objective in Equation~\ref{eq:ideal_map_learning} would require
optimizing a separate latent variable \(z_k\) for every training example. To
avoid this inner optimization, we amortize the latent explanation using a neural
encoder \(q_\phi\), which predicts
$z_k \simeq q_\phi(x_k).$
Substituting this amortized estimate into the MAP objective gives the practical
training objective
\begin{equation}
\label{eq:amortized_map_learning}
\max_{\theta_p,\phi}
\sum_{(x,y)\in D}
\log p_{\theta_p}(x,y,q_\phi(x)).
\end{equation}
Expanding the joint score, the loss becomes
\begin{equation}
\label{eq:det_map_objective}
\max_{\theta_p,\phi}
\sum_{(x,y)\in D}
\Big[
\log p_{\theta_p}\big(
y|
\mathbf{s}^\star(q_\phi(x)),
\boldsymbol{\beta}^\star(q_\phi(x))
\big)
+
\log p_{\theta_p}\big(
x|
\mathbf{s}^\star(q_\phi(x)),
\boldsymbol{\beta}^\star(q_\phi(x))
\big)
+
\log p\big(q_\phi(x)\big)
\Big]
\end{equation}

The first term encourages the inferred object-centric representation to explain
the symbolic task label. The second term encourages the same representation to
reconstruct the input image. The third term is a prior term on the amortized
latent explanation. Thus, the encoder is trained to produce a single latent
explanation that jointly accounts for the image, the task label, and the prior.

\section{Experiments} \label{sec:experiments}

We focus on evaluating the generalization capabilities of \our{}, as generalization under distributional shifts offers the clearest evidence that a model is truly learning to decompose scenes into meaningful components and reason over them \citep{Greff2020, Dittadi2021}. 
We pose the following research questions:


\noindent \textbf{Q1: Compositional generalization:} Can the model handle novel combinations of objects seen during training?\newline
\noindent \textbf{Q2: Task generalization:} Can the model generalize to new tasks not present in the training data?\newline
\noindent \textbf{Q3: Object count generalization:} Can the model handle 
object configurations larger (extrapolation) or smaller (interpolation) than those seen during training? \newline
\textbf{Q4: Stronger learning signal:}
Does our method provide a stronger object-level learning signal than purely neural task prediction? \newline
\textbf{Q5: Foundation models}: How do pretrained Vision-Language Models (VLMs) compare to our approach on structured reasoning tasks?

\subsection{Setup}

\textbf{Datasets}. We consider 4 datasets. For additional details, we refer to Appendix \ref{ap:expdetails}. \textit{(1) MultiMNIST-Addition} (MM-A) is our controlled diagnostic for learning object detection and predicates from task labels only. Each image contains up to three MNIST digits, and the only training label is their sum; digit identities, locations, and object counts are not used for training. We evaluate three shifts: compositional generalization, by training on 75\% of digit combinations with 0--3 digits and testing on the remaining 25\%; extrapolation, by testing on 4- and 5-digit images with either seen or unseen sums; and interpolation, by training on 0-, 1-, and 3-digit images and testing on 2-digit images.
\textit{(2) PokerRules}
tests whether learned object predicates can support relational rules beyond digit arithmetic. Images contain multiple playing cards and labels are poker-hand categories defined by symbolic relations among card ranks. We evaluate supplied rules for hand types not seen during training and extrapolation to scenes with more cards. This setting also probes an important limitation: some task labels identify relational structure, such as equality of ranks, more strongly than human-named card classes.
\textit{(3) CLEVR-Addition (CLEVR-A)} tests objectness and counting in scenes with more varied visual structure. We form pairs of CLEVR images and label each pair by the total number of objects across the two images. The evaluation includes object-count extrapolation, including seen and unseen output sums.
With \textit{(4) COCOLogic}  \citep{steinmann2025object} we study how task-level logical supervision interacts with partial object-level annotations in a task with more natural images.

\textbf{Baselines.}
We compare our approach against four neural baselines: a standard CNN, the Slot Attention model, MESH \citep{zhang2023unlocking}, and CoSA \citep{kori2023grounded}. Notably, CoSA is a recent work and, to the best of our knowledge, the only existing method that has explicitly attempted tasks of this type. We further include a Transformer predictor (TF), which replaces the MLP over slots with a Transformer encoder. Additionally, for tasks that do not require generalization to larger numbers of objects, we include comparisons with a standard neurosymbolic (NeSy) implementation following the setup described in \citet{misino2022vael} for the classification task. For working with real-world images, we additionally include DINOSAUR \citep{seitzer2022bridging}. In the experiment with COCOLogic we use DINOSAUR both as a competitor and as a backbone for our method. Full details of the baselines and implementations are provided in Appendix \ref{ap:expdetails}.

\textbf{Metrics.} 
We report task accuracy, defined as the proportion of correctly classified images, and concept accuracy, which measures the correctness of the inferred object representations. Note that concept accuracy is not computable for fully neural baselines, as they do not model  objects explicitly. In the COCOLogic setting, we report balanced accuracy, as done in the original paper \citep{steinmann2025object}.

\section{Results}
We evaluate \our's ability to generalize across the four core research questions introduced above. 
Additional experiments and ablation studies are provided in Appendix \ref{ap:additional}.

\textbf{Q1: \our{} generalizes more robustly to novel combinations of familiar objects.} In the MM-A task (Table \ref{tab:multimnist}), our model achieves a Task Accuracy of 90\% on out-of-distribution (OOD) compositions, higher than the neural and object-centric baselines. This demonstrates the model’s ability to exploit the provided logic knowledge to learn task-relevant object predicates that are more independent on the particular combinations seen in the training set.  
The comparison with NeSy helps isolate the role of object-centric perception: although NeSy uses the same symbolic reasoning component, it achieves only 35.76\% in this setting.


\textbf{Q2: \our{} allows for generalization to novel classes.} In the PokerRules task (Table \ref{tab:pokerrules}), our model achieves an OOD class accuracy of 72\%, compared with 0.46\% for NeSy; neural baselines are not directly applicable in this setting because their output spaces are fixed by the training labels.  Similar results are also reported for the MM-A and the CLEVR-A tasks (see Appendix \ref{ap:additional}). These results are consistent with the model learning object-level predicates that can be reused by new supplied rules, rather than only fitting the original image-level labels, as the Concept Accuracy results shows in Table \ref{tab:multimnist}.
This object-centric interface makes it possible to apply new symbolic rules over the learned predicates: by grounding predictions in symbolic concepts, the same perceptual model \textit{can be reused for different reasoning tasks on the same input domain} simply by modifying the logical program, with performance depending on how well the learned predicates align with the concepts required by the new rule. 


\begin{table*}[t]
\centering
\caption{Task Accuracy (overall image) and Concept Accuracy (object-level decomposition) in MM-A. Results are on a test set of in-distribution compositions (\textbf{Test}) and on out-of-distribution compositions (\textbf{OOD}) . We also report extrapolation performance on images with four or five digits OOD counts. 
Dashes indicate cases where a model cannot provide results due to architectural limitations.}

\begin{tabular}{lcccccc}
\toprule
\multicolumn{1}{c}{MM-A} & \multicolumn{2}{c}{Task Acc.} & \multicolumn{2}{c}{Concept Acc.} & \multicolumn{2}{c}{Extrapolation} \\
\cmidrule(lr){2-3} \cmidrule(lr){4-5} \cmidrule(lr){6-7}  & Test & OOD & Test & OOD & 4 digits & 5 digits \\
\midrule
CNN & 79.90${\scriptstyle \pm 0.17}$ & 3.43${\scriptstyle \pm 0.15}$ & \unavailable & \unavailable & 22.16${\scriptstyle \pm 0.90}$ & 13.16${\scriptstyle \pm 0.96}$  \\
SA & \textbf{98.90${\scriptstyle \pm 0.34}$} & 13.30${\scriptstyle \pm 3.14}$ & \unavailable & \unavailable & 36.23${\scriptstyle \pm 4.44}$ & 9.76${\scriptstyle \pm 6.11}$ \\
MESH & 98.86${\scriptstyle \pm 0.47}$ & 18.26${\scriptstyle \pm 1.33}$ & \unavailable & \unavailable & 37.50${\scriptstyle \pm 1.15}$ & 12.00${\scriptstyle \pm 0.51}$\\
Tf & 95.20${\scriptstyle \pm 0.42}$ & 52.28${\scriptstyle \pm 9.11}$ & \unavailable & \unavailable & 50.09${\scriptstyle \pm 8.12}$ & 18.42${\scriptstyle \pm 9.85}$ \\
CoSA & 93.00${\scriptstyle \pm 1.92}$ & 36.2${\scriptstyle \pm 15.10}$ & \unavailable & \unavailable & 52.20${\scriptstyle \pm 14.01}$ & 25.06${\scriptstyle \pm 12.82}$ \\
NeSy & 90.70${\scriptstyle \pm 4.02}$ & 35.76${\scriptstyle \pm 6.24}$ & 55.43${\scriptstyle \pm 26.40}$ & 23.83${\scriptstyle \pm 3.82}$ & \unavailable & \unavailable \\
\textbf{Ours} & 94.26${\scriptstyle \pm 2.00}$ & \textbf{90.00${\scriptstyle \pm 3.01}$} & \textbf{85.16${\scriptstyle \pm 2.45}$} & \textbf{65.46${\scriptstyle \pm 5.70}$} & \textbf{69.73${\scriptstyle \pm 10.74}$} & \textbf{44.06${\scriptstyle \pm 3.85}$}\\
\bottomrule
\end{tabular}
\label{tab:multimnist}
\end{table*}
\begin{table*}[t]
\centering
\caption{Task Accuracy on PokerRules for both in-distribution test classes and OOD classes. A dash indicates cases where a model cannot provide results due to architectural limitations.}
\begin{tabular}{lcccc}
\toprule
 & Test & OOD class & In-distribution class & OOD class\\
 \midrule
\multicolumn{1}{c}{PokerRules} & \multicolumn{2}{c}{} & \multicolumn{2}{c}{Extrapolation: 5 cards}\\
\cmidrule(lr){2-3} \cmidrule(lr){4-5}
CNN & 81.96${\scriptstyle \pm 1.10}$ & \unavailable & 22.03${\scriptstyle \pm 5.27}$ & \unavailable \\
SA & 99.30${\scriptstyle \pm 1.21}$ & \unavailable & 35.86${\scriptstyle \pm 8.72}$ & \unavailable \\
MESH & \textbf{99.93}${\scriptstyle \pm 0.11}$ & \unavailable & 37.80${\scriptstyle \pm 4.91}$ & \unavailable \\
Tf & 97.41${\scriptstyle \pm 0.15}$ & \unavailable & 15.20${\scriptstyle \pm 7.41}$ & \unavailable \\
CoSA & 95.46${\scriptstyle \pm 3.90}$ & \unavailable & 44.26${\scriptstyle \pm 17.16}$ & \unavailable \\
NeSy & 80.23${\scriptstyle \pm 2.11}$ & 0.46${\scriptstyle \pm 0.05}$ & \unavailable & \unavailable \\
\textbf{Ours} & 97.90${\scriptstyle \pm 1.17}$ & \textbf{72.23}${\scriptstyle \pm 16.72}$ & \textbf{78.53}${\scriptstyle \pm 4.68}$ & \textbf{78.23}${\scriptstyle \pm 19.24}$ \\
\bottomrule
\end{tabular}
\label{tab:pokerrules}
\end{table*}

\textbf{Q3: \our{} adapts to new structural configurations.} When evaluated on images containing a number of objects not seen during training, our model generalizes more effectively than neural baselines. In the MM-A task, the model maintains substantially higher accuracy on extrapolation test sets with four and five digits, while competitors show significant degradation. The interpolation results, reported in Appendix \ref{ap:additional}, also show a similar trend. Similarly, in the PokerRules and CLEVR-A tasks, the model 
shows stronger extrapolation than competitors both on known classes and previously unseen class configurations, where other approaches either degrade substantially or are not applicable. 
This result is consistent with the intended role of objectness: slots can be activated or suppressed depending on the input, allowing the same symbolic rule to operate over different numbers of predicted objects.

\begin{figure}[t]
    \centering
    \begin{minipage}[t]{0.48\linewidth}
        \centering
        \includegraphics[width=\linewidth]{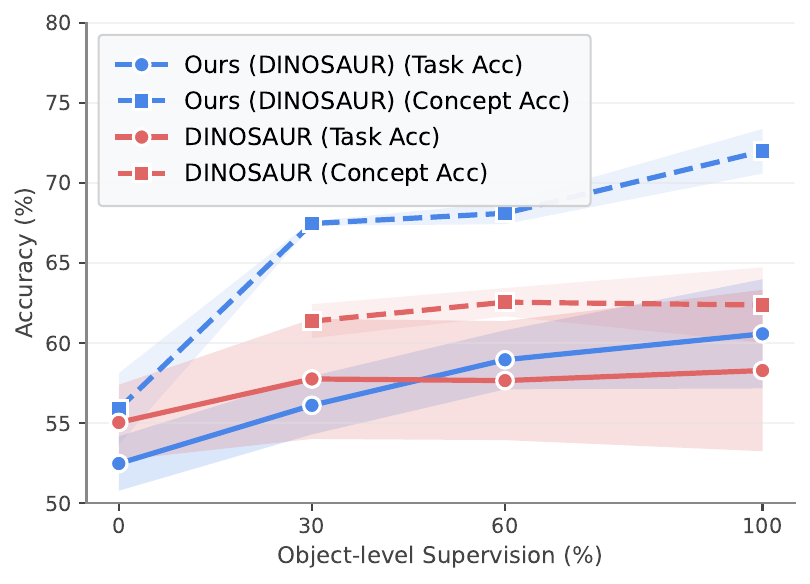}
        \caption{Concept and task accuracy on COCOLogic under varying levels of object-level supervision.}
        \label{fig:sup_acc}
    \end{minipage}
    \hfill
    \begin{minipage}[t]{0.48\linewidth}
        \centering
        \includegraphics[width=0.9\linewidth]{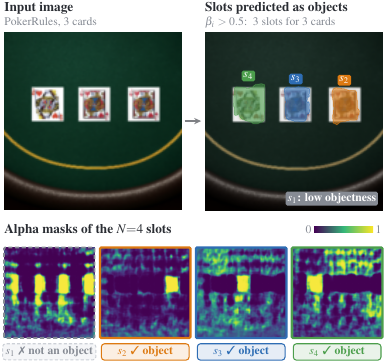}  
        \caption{Alpha masks of the slots learned by DeepObjectLog on PokerRules. Slots that attend to a single card are predicted as objects ($\beta_i > 0.5$), while the slot spread over multiple regions receives low objectness.}
        \label{fig:alphamask_main}
    \end{minipage}
\end{figure}
\paragraph{Q4: Logical reasoning provides a stronger task-level signal for object-centric learning.} As shown in Figure~\ref{fig:sup_acc}, on COCOLogic, at 0\% object-level supervision DINOSAUR cannot provide concept predictions, while our method already achieves non-trivial concept accuracy, demonstrating its ability to learn object structure purely from the task signal. This advantage is maintained as the level of supervision increases. Task accuracy remains comparable across methods; notably, this is achieved while preserving the interpretability and guarantees of our logical reasoning component. For reference, a Concept Bottleneck Model in~\cite{steinmann2025object} achieves a task accuracy of 58.84, which is in line with our results.

\textbf{Q5: \our{} compares favorably to pretrained Vision-Language Models on MM-A.}
We evaluated LLaVA-1.5-7B-hf and Qwen2.5-VL-7B-Instruct on MM-A by providing the task description and possible answers. Under this prompting setup, both models struggled: the best competitor achieved only 59.93\% accuracy on OOD compositions. Under the same benchmark evaluation, \our{} reached 90\% accuracy. While this comparison is not strictly fair, since these models are not explicitly trained for this task, it demonstrates that the benchmark is non-trivial even for strong general-purpose VLMs. Details and results are provided in Appendix \ref{ap:additional}.

\section{Conclusions} \label{sec:conclusions}
We propose \our{}, a neurosymbolic model that integrates object-centric perception with probabilistic logical reasoning to jointly infer and reason over a variable number of latent objects. Our results show that (1) the proposed method strongly outperforms neural and neurosymbolic baselines in compositional generalization tasks, (2) it effectively generalizes to scenes with previously unseen object counts and classes, and (3) it enables robust reasoning under uncertainty without requiring explicit object-level supervision. These advances mark an important step toward more general neurosymbolic systems, contributing to the development of robust and interpretable AI, though future deployments must mitigate risks of bias and dual-use surveillance.


\textbf{Limitations and future work.} Our approach shares limitations with current object-centric and neurosymbolic methods. First, exact inference makes the model less scalable than purely neural approaches. While we mitigate this using \cite{maene2024klay} (Appendix \ref{ap:additional}), approximate methods could be adopted to further address this limitation. 
Second, weakly supervised models often fall back on reasoning shortcuts \cite{marconato2023not}, but our contribution is a step toward overcoming these challenges by enforcing structural constraints that discourage such behaviors.

\section*{Acknowledgements}
The authors would like to express their gratitude to Luc De Raedt and Robin Manhaeve for the early discussions on the subject of this research.
This research has received funding from the KU Leuven Research Fund (GA No. STG/22/021, CELSA/24/008), from the Research Foundation-Flanders FWO (GA No. 1185125N, G047124N),  from the Flemish Government under the “Onderzoeksprogramma Artificiële Intelligentie (AI) Vlaanderen” programme and from the European Research Council (ERC) under the European Union’s Horizon 2020 research and innovation programme (GA No. 101142702).

\bibliographystyle{abbrvnat}
\bibliography{bibliography}

@article{marra2024statistical,
  title={From statistical relational to neurosymbolic artificial intelligence: A survey},
  author={Marra, Giuseppe and Duman{\v{c}}i{\'c}, Sebastijan and Manhaeve, Robin and De Raedt, Luc},
  journal={Artificial Intelligence},
  volume={328},
  pages={104062},
  year={2024},
  publisher={Elsevier}
}

@article{manhaeve2018deepproblog,
  title={Deepproblog: Neural probabilistic logic programming},
  author={Manhaeve, Robin and Dumancic, Sebastijan and Kimmig, Angelika and Demeester, Thomas and De Raedt, Luc},
  journal={Advances in neural information processing systems},
  volume={31},
  year={2018}
}

@inproceedings{de2023neural,
  title={Neural probabilistic logic programming in discrete-continuous domains},
  author={De Smet, Lennert and Dos Martires, Pedro Zuidberg and Manhaeve, Robin and Marra, Giuseppe and Kimmig, Angelika and De Readt, Luc},
  booktitle={Uncertainty in Artificial Intelligence},
  pages={529--538},
  year={2023},
  organization={PMLR}
}

@article{misino2022vael,
  title={Vael: Bridging variational autoencoders and probabilistic logic programming},
  author={Misino, Eleonora and Marra, Giuseppe and Sansone, Emanuele},
  journal={Advances in Neural Information Processing Systems},
  volume={35},
  pages={4667--4679},
  year={2022}
}

@inproceedings{skryagin2022neural,
  title={Neural-probabilistic answer set programming},
  author={Skryagin, Arseny and Stammer, Wolfgang and Ochs, Daniel and Dhami, Devendra Singh and Kersting, Kristian},
  booktitle={Proceedings of the International Conference on Principles of Knowledge Representation and Reasoning},
  volume={19},
  number={1},
  pages={463--473},
  year={2022}
}

@article{mao2019neuro,
  title={The neuro-symbolic concept learner: Interpreting scenes, words, and sentences from natural supervision},
  author={Mao, Jiayuan and Gan, Chuang and Kohli, Pushmeet and Tenenbaum, Joshua B and Wu, Jiajun},
  journal={arXiv preprint arXiv:1904.12584},
  year={2019}
}

@inproceedings{stammer2021right,
  title={Right for the right concept: Revising neuro-symbolic concepts by interacting with their explanations},
  author={Stammer, Wolfgang and Schramowski, Patrick and Kersting, Kristian},
  booktitle={Proceedings of the IEEE/CVF conference on computer vision and pattern recognition},
  pages={3619--3629},
  year={2021}
}

@article{shindo2023alpha,
  title={$\alpha$ ilp: thinking visual scenes as differentiable logic programs},
  author={Shindo, Hikaru and Pfanschilling, Viktor and Dhami, Devendra Singh and Kersting, Kristian},
  journal={Machine Learning},
  volume={112},
  number={5},
  pages={1465--1497},
  year={2023},
  publisher={Springer}
}

@article{yi2018neural,
  title={Neural-symbolic vqa: Disentangling reasoning from vision and language understanding},
  author={Yi, Kexin and Wu, Jiajun and Gan, Chuang and Torralba, Antonio and Kohli, Pushmeet and Tenenbaum, Josh},
  journal={Advances in neural information processing systems},
  volume={31},
  year={2018}
}

@article{de2024object,
  title={Object-centric learning with capsule networks: A survey},
  author={De Sousa Ribeiro, Fabio and Duarte, Kevin and Everett, Miles and Leontidis, Georgios and Shah, Mubarak},
  journal={ACM Computing Surveys},
  volume={56},
  number={11},
  pages={1--291},
  year={2024},
  publisher={ACM New York, NY}
}

@article{locatello2020object,
  title={Object-centric learning with slot attention},
  author={Locatello, Francesco and Weissenborn, Dirk and Unterthiner, Thomas and Mahendran, Aravindh and Heigold, Georg and Uszkoreit, Jakob and Dosovitskiy, Alexey and Kipf, Thomas},
  journal={Advances in neural information processing systems},
  volume={33},
  pages={11525--11538},
  year={2020}
}

@article{burgess2019monet,
  title={Monet: Unsupervised scene decomposition and representation},
  author={Burgess, Christopher P and Matthey, Loic and Watters, Nicholas and Kabra, Rishabh and Higgins, Irina and Botvinick, Matt and Lerchner, Alexander},
  journal={arXiv preprint arXiv:1901.11390},
  year={2019}
}

@inproceedings{greff2019multi,
  title={Multi-object representation learning with iterative variational inference},
  author={Greff, Klaus and Kaufman, Rapha{\"e}l Lopez and Kabra, Rishabh and Watters, Nick and Burgess, Christopher and Zoran, Daniel and Matthey, Loic and Botvinick, Matthew and Lerchner, Alexander},
  booktitle={International conference on machine learning},
  pages={2424--2433},
  year={2019},
  organization={PMLR}
}

@article{engelcke2019genesis,
  title={Genesis: Generative scene inference and sampling with object-centric latent representations},
  author={Engelcke, Martin and Kosiorek, Adam R and Jones, Oiwi Parker and Posner, Ingmar},
  journal={arXiv preprint arXiv:1907.13052},
  year={2019}
}

@article{engelcke2021genesisv2,
  title={Genesis-v2: Inferring unordered object representations without iterative refinement},
  author={Engelcke, Martin and Parker Jones, Oiwi and Posner, Ingmar},
  journal={Advances in Neural Information Processing Systems},
  volume={34},
  pages={8085--8094},
  year={2021}
}

@article{elsayed2022savi++,
  title={Savi++: Towards end-to-end object-centric learning from real-world videos},
  author={Elsayed, Gamaleldin and Mahendran, Aravindh and Van Steenkiste, Sjoerd and Greff, Klaus and Mozer, Michael C and Kipf, Thomas},
  journal={Advances in Neural Information Processing Systems},
  volume={35},
  pages={28940--28954},
  year={2022}
}

@article{singh2022simple,
  title={Simple unsupervised object-centric learning for complex and naturalistic videos},
  author={Singh, Gautam and Wu, Yi-Fu and Ahn, Sungjin},
  journal={Advances in Neural Information Processing Systems},
  volume={35},
  pages={18181--18196},
  year={2022}
}

@article{seitzer2022bridging,
  title={Bridging the gap to real-world object-centric learning},
  author={Seitzer, Maximilian and Horn, Max and Zadaianchuk, Andrii and Zietlow, Dominik and Xiao, Tianjun and Simon-Gabriel, Carl-Johann and He, Tong and Zhang, Zheng and Sch{\"o}lkopf, Bernhard and Brox, Thomas and others},
  journal={arXiv preprint arXiv:2209.14860},
  year={2022}
}

@inproceedings{zhang2023unlocking,
  title={Unlocking slot attention by changing optimal transport costs},
  author={Zhang, Yan and Zhang, David W and Lacoste-Julien, Simon and Burghouts, Gertjan J and Snoek, Cees GM},
  booktitle={International Conference on Machine Learning},
  pages={41931--41951},
  year={2023},
  organization={PMLR}
}

@inproceedings{kori2023grounded,
  title={Grounded Object-Centric Learning},
  author={Kori, Avinash and Locatello, Francesco and Ribeiro, Fabio De Sousa and Toni, Francesca and Glocker, Ben},
  booktitle={ICLR},
  year={2024}
}

@inproceedings{fan2024adaptive,
  title={Adaptive slot attention: Object discovery with dynamic slot number},
  author={Fan, Ke and Bai, Zechen and Xiao, Tianjun and He, Tong and Horn, Max and Fu, Yanwei and Locatello, Francesco and Zhang, Zheng},
  booktitle={Proceedings of the IEEE/CVF Conference on Computer Vision and Pattern Recognition},
  pages={23062--23071},
  year={2024}
}

@article{marconato2023not,
  title={Not all neuro-symbolic concepts are created equal: Analysis and mitigation of reasoning shortcuts},
  author={Marconato, Emanuele and Teso, Stefano and Vergari, Antonio and Passerini, Andrea},
  journal={Advances in Neural Information Processing Systems},
  volume={36},
  pages={72507--72539},
  year={2023}
}

@article{wiedemer2023provable,
  title={Provable compositional generalization for object-centric learning},
  author={Wiedemer, Thadd{\"a}us and Brady, Jack and Panfilov, Alexander and Juhos, Attila and Bethge, Matthias and Brendel, Wieland},
  journal={arXiv preprint arXiv:2310.05327},
  year={2023}
}

@article{Greff2020,
   author = {Klaus Greff and Sjoerd van Steenkiste and Jürgen Schmidhuber},
   month = {12},
   title = {On the Binding Problem in Artificial Neural Networks},
   year = {2020}
}

@article{Dittadi2021,
   author = {Andrea Dittadi and Samuele Papa and Michele De Vita and Bernhard Schölkopf and Ole Winther and Francesco Locatello},
   month = {7},
   title = {Generalization and Robustness Implications in Object-Centric Learning},
   year = {2021}
}

@article{kori2024identifiable,
  title={Identifiable object-centric representation learning via probabilistic slot attention},
  author={Kori, Avinash and Locatello, Francesco and Santhirasekaram, Ainkaran and Toni, Francesca and Glocker, Ben and De Sousa Ribeiro, Fabio},
  journal={Advances in Neural Information Processing Systems},
  volume={37},
  pages={93300--93335},
  year={2024}
}

@article{maene2024klay,
  title={KLay: Accelerating Neurosymbolic AI},
  author={Maene, Jaron and Derkinderen, Vincent and Martires, Pedro Zuidberg Dos},
  journal={arXiv preprint arXiv:2410.11415},
  year={2024}
}

@inproceedings{de2007problog,
  title={ProbLog: A probabilistic Prolog and its application in link discovery},
  author={De Raedt, Luc and Kimmig, Angelika and Toivonen, Hannu},
  booktitle={IJCAI 2007, Proceedings of the 20th international joint conference on artificial intelligence},
  pages={2462--2467},
  year={2007},
  organization={IJCAI-INT JOINT CONF ARTIF INTELL}
}

@inproceedings{gao2024knowledge,
  title={Knowledge-enhanced historical document segmentation and recognition},
  author={Gao, En-Hao and Huang, Yu-Xuan and Hu, Wen-Chao and Zhu, Xin-Hao and Dai, Wang-Zhou},
  booktitle={Proceedings of the AAAI Conference on Artificial Intelligence},
  volume={38},
  number={8},
  pages={8409--8416},
  year={2024}
}

@inproceedings{cai2021abductive,
  title={Abductive Learning with Ground Knowledge Base.},
  author={Cai, Le-Wen and Dai, Wang-Zhou and Huang, Yu-Xuan and Li, Yufeng and Muggleton, Stephen H and Jiang, Yuan},
  booktitle={IJCAI},
  pages={1815--1821},
  year={2021}
}

@inproceedings{johnson2017clevr,
  title={Clevr: A diagnostic dataset for compositional language and elementary visual reasoning},
  author={Johnson, Justin and Hariharan, Bharath and Van Der Maaten, Laurens and Fei-Fei, Li and Lawrence Zitnick, C and Girshick, Ross},
  booktitle={Proceedings of the IEEE conference on computer vision and pattern recognition},
  pages={2901--2910},
  year={2017}
}

@misc{mulitdigitmnist,
  author = {Sun, Shao-Hua},
  title = {Multi-digit MNIST for Few-shot Learning},
  year = {2019},
  journal = {GitHub repository},
  url = {https://github.com/shaohua0116/MultiDigitMNIST},
}

@article{shindo2024deisam,
  title={Deisam: Segment anything with deictic prompting},
  author={Shindo, Hikaru and Brack, Manuel and Sudhakaran, Gopika and Dhami, Devendra S and Schramowski, Patrick and Kersting, Kristian},
  journal={Advances in Neural Information Processing Systems},
  volume={37},
  pages={52266--52295},
  year={2024}
}

@article{wu2022slotformer,
  title={Slotformer: Unsupervised visual dynamics simulation with object-centric models},
  author={Wu, Ziyi and Dvornik, Nikita and Greff, Klaus and Kipf, Thomas and Garg, Animesh},
  journal={arXiv preprint arXiv:2210.05861},
  year={2022}
}

@article{steinmann2025object,
  title={Object centric concept bottlenecks},
  author={Steinmann, David and Stammer, Wolfgang and W{\"u}st, Antonia and Kersting, Kristian},
  journal={arXiv preprint arXiv:2505.24492},
  year={2025}
}

@article{manhaeve2026deeplog,
  title={DeepLog: A Software Framework for Modular Neurosymbolic AI},
  author={Manhaeve, Robin and Colamonaco, Stefano and Derkinderen, Vincent and Adriaensen, Rik and Van Praet, Lucas and De Raedt, Luc and Marra, Giuseppe},
  journal={arXiv preprint arXiv:2605.10279},
  year={2026}
}

@article{van2023nesi,
  title={A-nesi: A scalable approximate method for probabilistic neurosymbolic inference},
  author={van Krieken, Emile and Thanapalasingam, Thiviyan and Tomczak, Jakub and Van Harmelen, Frank and Ten Teije, Annette},
  journal={Advances in Neural Information Processing Systems},
  volume={36},
  pages={24586--24609},
  year={2023}
}

\newpage

\appendix
{\huge \textbf{Supplementary Material}}

\addtocontents{toc}{\protect\setcounter{tocdepth}{2}}
\renewcommand{\contentsname}{\large Table of Contents}
\tableofcontents
\newpage


\section{Amortized MAP objective}
\label{app:map}

The learning objective used in \our{} follows a deterministic MAP principle.
For each image--label pair \((x,y)\), we seek a single latent explanation \(z\)
that receives high probability under the model. The ideal objective is therefore
\begin{equation}
\label{eq:app_ideal_map}
\mathcal{J}_{\mathrm{MAP}}(\theta_p)
=
\sum_{(x,y)\in D}
\max_z
\log p_{\theta_p}(x,y,z).
\end{equation}
Equivalently, if we introduce one latent variable \(z_k\) for each training
example \((x_k,y_k)\), this can be written as
\begin{equation}
\label{eq:app_latent_optimization}
\max_{\theta_p,\{z_k\}_{k=1}^K}
\sum_{k=1}^K
\log p_{\theta_p}(x_k,y_k,z_k).
\end{equation}

We now derive the form of the score \(p_{\theta_p}(x,y,z)\). Starting from the
full joint model,
\begin{equation}
\label{eq:app_full_joint}
p_{\theta_p}
\big(
y,x,\mathbf{s},\boldsymbol{\beta},\mathbf{o},\mathbf{c},z
\big),
\end{equation}
the score of a candidate latent explanation \(z\) is obtained by marginalizing
the slot variables \(\mathbf{s}\), the objectness scores
\(\boldsymbol{\beta}\), the objectness assignments \(\mathbf{o}\), and the class
assignments \(\mathbf{c}\):
\begin{equation}
\label{eq:app_score_marginal}
p_{\theta_p}(x,y,z)
=
\sum_{\mathbf{o},\mathbf{c}}
\int d\mathbf{s}\,d\boldsymbol{\beta}\;
p_{\theta_p}
\big(
y,x,\mathbf{s},\boldsymbol{\beta},\mathbf{o},\mathbf{c},z
\big).
\end{equation}

In our deterministic setting, the conditional distributions over
\(\mathbf{s}\) and \(\boldsymbol{\beta}\) are Dirac distributions centered at
the outputs of the neural modules:
\[
\mathbf{s} = \mathbf{s}^\star(z),
\qquad
\boldsymbol{\beta} = \boldsymbol{\beta}^\star(z).
\]
Using the sifting property of Dirac distributions, these variables collapse out
of Equation~\ref{eq:app_score_marginal}. The remaining discrete variables are
marginalized by the probabilistic logic layer. This yields
\begin{equation}
\label{eq:app_collapsed_score}
p_{\theta_p}(x,y,z)
=
p(z)\,
p_{\theta_p}
\big(
x\mid
\mathbf{s}^\star(z),
\boldsymbol{\beta}^\star(z)
\big)\,
p_{\theta_p}
\big(
y\mid
\mathbf{s}^\star(z),
\boldsymbol{\beta}^\star(z)
\big).
\end{equation}

The task likelihood is computed by summing over the discrete objectness and
class variables:
\begin{equation}
\label{eq:app_task_likelihood}
p_{\theta_p}
\big(
y\mid \mathbf{s},\boldsymbol{\beta}
\big)
=
\sum_{\mathbf{o},\mathbf{c}}
p_L(y\mid \mathbf{o},\mathbf{c})
\prod_i
p(o_i\mid \beta_i)
p_{\theta_p}(c_i\mid s_i,\beta_i).
\end{equation}
Thus, although the continuous part of the model is treated deterministically, the
symbolic prediction remains probabilistic and marginalizes over possible
objectness and class assignments.

Taking logs in Equation~\ref{eq:app_collapsed_score}, the score of a latent
explanation becomes
\begin{equation}
\label{eq:app_log_score}
\log p_{\theta_p}(x,y,z)
=
\log p_{\theta_p}
\big(
y\mid
\mathbf{s}^\star(z),
\boldsymbol{\beta}^\star(z)
\big)
+
\log p_{\theta_p}
\big(
x\mid
\mathbf{s}^\star(z),
\boldsymbol{\beta}^\star(z)
\big)
+
\log p(z).
\end{equation}
This is the log joint density of the observed image, the observed label, and the
candidate latent explanation. For fixed \(x\) and \(y\), maximizing this score
with respect to \(z\) is equivalent to MAP inference:
\begin{equation}
\label{eq:app_map_equivalence}
\arg\max_z p_{\theta_p}(z\mid x,y)
=
\arg\max_z p_{\theta_p}(x,y,z),
\end{equation}
because
\[
p_{\theta_p}(z\mid x,y)
=
\frac{
p_{\theta_p}(x,y,z)
}{
p_{\theta_p}(x,y)
},
\]
and the denominator does not depend on \(z\).

However, solving this MAP problem independently for every training example would
require an inner optimization loop. We therefore amortize this step with a
deterministic encoder \(q_\phi\), replacing the per-example MAP variable by
\[
z \simeq q_\phi(x).
\]
The resulting amortized MAP objective is
\begin{equation}
\label{eq:app_amortized_map}
\mathcal{J}_{\mathrm{AMAP}}(\theta_p,\phi)
=
\sum_{(x,y)\in D}
\log p_{\theta_p}(x,y,q_\phi(x)).
\end{equation}
Expanding the score gives 
\begin{align}
\label{eq:app_expanded_amortized_map}
\mathcal{J}_{\mathrm{AMAP}}(\theta_p,\phi)
=
\sum_{(x,y)\in D}
\Big[
\log p_{\theta_p}
\big(
y\mid
\mathbf{s}^\star(q_\phi(x)),
\boldsymbol{\beta}^\star(q_\phi(x))
\big) \notag \\  \notag
+
\log p_{\theta_p}
\big(
x\mid
\mathbf{s}^\star(q_\phi(x)),
\boldsymbol{\beta}^\star(q_\phi(x))
\big)\\
+
\log p(q_\phi(x))
\Big]
\end{align}

For fixed model parameters, the amortized objective is bounded above by the
ideal per-example MAP objective:
\begin{equation}
\label{eq:app_amortization_gap}
\log p_{\theta_p}(x,y,q_\phi(x))
\leq
\max_z
\log p_{\theta_p}(x,y,z).
\end{equation}

\subsection{Prior and Activation Regularization}

The term \(\log p(q_\phi(x))\) is the prior evaluated at the amortized latent
explanation. It regularizes the encoder output.

Assuming a standard isotropic Gaussian prior,
\[
p(z)=\mathcal{N}(0,I),
\]
we have
\[
\log p(z)
=
-\frac{1}{2}\|z\|^2
+
\mathrm{const}.
\]
Substituting \(z=q_\phi(x)\) gives
\[
\log p(q_\phi(x))
=
-\frac{1}{2}\|q_\phi(x)\|^2
+
\mathrm{const}.
\]
Therefore, maximizing the prior term is equivalent, up to an additive constant,
to minimizing
\[
\frac{1}{2}\|q_\phi(x)\|^2.
\]
We refer to this as activation regularization because it penalizes the magnitude
of the encoder output rather than the model parameters.

\section{Experimental details} \label{ap:expdetails}

\subsection{Architectural details}
All architectures and code used in this study were implemented in Python using PyTorch, with DeepLog \citep{manhaeve2026deeplog} employed for the symbolic integration. We provide here the architectural details of the baseline models and our proposed method used in the experiments.
\paragraph{CNN}
The CNN baseline consists of a convolutional encoder followed by a multi-layer perceptron (MLP) classifier over the target classes. The encoder comprises a sequence of convolutional layers with kernel size 5, each followed by a ReLU activation. The first convolutional layer uses a stride of 2 to reduce spatial resolution, while the remaining layers use a stride of 1. The output of the final convolutional layer is processed by an Adaptive Average Pooling layer. The resulting feature vector is then passed to an MLP with hidden dimension 64, which outputs the class logits. 

\paragraph{Slot Attention}
For Slot Attention, we adopt the architecture from the original paper. Both the slot dimension and the embedding dimension are set to 64, and the number of slot attention iterations is 2. For the final classification, we apply an MLP over the summed slot embeddings. The decoder module follows the design proposed in the original Slot Attention paper and several PyTorch implementations are available online \footnote{https://github.com/amazon-science/AdaSlot}.

\paragraph{MESH}
The MESH baseline builds on the Slot Attention architecture, with the addition of mesh-based interactions \citep{zhang2023unlocking} \footnote{https://github.com/davzha/MESH}. Both the slot dimension and the embedding dimension are set to 64, and the number of slot attention iterations is 2. The remaining hyperparameters match those in the original MESH repository. For the final classification, as in Slot Attention, we apply an MLP over the summed slot embeddings. The decoder employed matches that used in the Slot Attention baseline.

\paragraph{Transformer predictor (Tf)} The Transformer predictor shares the encoder of the MESH baseline, but replaces the MLP over the summed slot embeddings with a Transformer encoder that attends over all slots. Each slot is linearly projected to dimension 128, and a learnable [CLS] token is prepended to the sequence. We use no positional embeddings, so the predictor is invariant to the order of the slots and accepts any number of them. The sequence is processed by 3 pre-norm Transformer encoder layers with 4 attention heads, feed-forward dimension 256, GELU activations and dropout 0.1. The output of the [CLS] token is layer-normalized and passed to an MLP with hidden dimension 128, which outputs the class logits.

\paragraph{CoSA}
The CoSA implementation code is provided by the authors of the paper and is available online, even if not all hyperparameters of the model were explicitly provided in the released code \footnote{https://github.com/koriavinash1/CoSA}. Based on reported results, we selected the Gumbel variant with 3 slot attention iterations as the best-performing configuration and used it across all experiments. For the MM-A and Poker Rules tasks we adapted the reasoning classifier described in their work to fit the specific tasks considered in our evaluation (i.e. setting the correct amount of properties and classes).\\ This model required structural modifications for the CLEVR-A task due to the two input images used to calculate the sum. To adapt the model, we used the Slot Attention mechanism on each image separately and added the resulting slots to feed everything into a single classification head. As the reasoning classifier described in their paper cannot be used for this task without modification, we replaced it with a simple MLP classifier.

\paragraph{NeSy baseline}
Following the idea exposed in \citet{misino2022vael}, in particular for the Label Classification task, the NeSy baseline uses a single CNN encoder backbone to process the input and leverages the DeepProbLog interface, mirroring our own method for fair comparison. To enable classification over multiple entities, the architecture includes two MLPs per possible object in the training set—one for estimating objectness and one for classification. All hyperparameters of the encoder align with those used in the CNN baseline.
The symbolic part is interfaced through DeepLog\citep{manhaeve2026deeplog}, where each output of the objectness and classifiers represents a neural predicate. These predicates are then utilised in combination to construct the rules employed for the final classification of the task. An example pseudo-code for the MultiMNIST-addition task is available in Table \ref{tab:pred_multimnist}.

\paragraph{DINOSAUR} We include DINOSAUR \citep{seitzer2022bridging} as a strong object-centric baseline designed for real-world images. We re-implemented DINOSAUR and use the variant based on a pretrained DINO backbone to obtain high-quality visual features. For the purely neural baseline, we follow the same classification strategy adopted for Slot Attention, applying an MLP over the aggregated slot representations. In our method, instead, the extracted slots are directly integrated into the neurosymbolic pipeline.

\paragraph{Our model}
Our proposed method was implemented using both the standard Slot Attention mechanism and the MESH variant. As the latter showed slightly better average performance, we report results for the MESH-based implementation. The hyperparameters follow those recommended in the MESH paper, where maintaining a low number of slot iterations (2) and increasing the number of mesh iterations (5) proved beneficial. Unlike the NeSy baseline, our method employs only two MLPs in total (one for objectness and one for classification), which are applied separately to each extracted object representation. To condition both the class classifier and the reconstruction, we follow the Zero Slot Strategy from Adaptive Slot Attention: the objectness probability is multiplied with the slot before it is passed to the classifier and decoder (See the next subsection for details.). The decoder architecture is the same as used in Slot Attention and the symbolic interface is the same as described in the NeSy baseline.\\
As with CoSA, this model required some structural modifications for the CLEVR-A task due to the two input images. We simply adapted the model by using the Slot Attention mechanism separately on the two images. All classification results were then fed into the same logic circuit. Please note that, for this specific task, adding a classification head for the object properties is unnecessary, as the objectness head is sufficient.

\textbf{Soft masking}
Our implementation of soft masking draws inspiration from the Zero Slot strategy introduced in Adaptive Slot Attention \citep{fan2024adaptive}. In their formulation, each slot representation $s_i$ is modulated by its associated objectness score (our $\beta_i$):
\[\tilde{s_i} = s_i \cdot \beta_i \]
However, unlike their approach, we do not apply a Gumbel-Softmax to discretize the objectness scores.

\textbf{On the number of slots $N$.} In our architecture, the hyperparameter $N$ serves as an upper bound on the number of representable objects. During training, we set $N$ equal to the maximum number of objects present in the training scenes (e.g., $N=3$ for MultiMNIST-Addition). For extrapolation to 4 or 5 digits, we increase $N$ at inference time to provide the necessary capacity. 
Interestingly, while initializing with a very high $N$ expands the combinatorial search space, our end-to-end logical signal mitigates this: if $N=9$ but the task is a sum of 9, the logic program naturally prunes impossible configurations (e.g., two 9s). Furthermore, the perceptual module and symbolic prior collaborate to suppress spurious slots via the objectness variable. For example, a ``0'' in MultiMNIST does not influence the sum, yet our model correctly identifies it with high objectness and the correct class. The perceptual reconstruction ensures the ``0'' is recognized as an entity, while the logic ensures the other digits align with the required sum, proving that extrapolation is driven by genuine scene decomposition rather than slot over-provisioning. This is supported by our interpolation experiments (Table \ref{tab:interp}), where the model dynamically adapts to an unseen number of objects within a fixed $N$. 
Finally, to rigorously evaluate this synergy, our concept accuracy relies on an exact match (subset accuracy) metric: the model must correctly predict the objectness and class for \textit{every} slot in the image, otherwise the entire prediction scores zero.

\textbf{Experiment with object-level supervision.} In the experimental setting designed to evaluate varying levels of object-level supervision (i.e., the COCOLogic benchmark), we adapt the training objective for both our proposed method and the DINOSAUR baseline by utilizing a combined loss function. Specifically, the total loss is formulated as the sum of the primary task-level loss, the reconstruction loss and an auxiliary object-level loss. Because the object-centric slots are inherently unordered and permutation-invariant, following common practices, we compute the object-level loss by first applying the Hungarian algorithm to establish an optimal bipartite matching between the predicted slots and the available ground-truth object labels. Once the optimal assignment is determined, a standard classification loss (e.g., cross-entropy) is applied to the matched slots. This symmetric formulation ensures that both our method and the baseline can effectively leverage partial explicit object annotations while continuing to rely on the global task loss for the overarching reasoning objective.

\subsection{Training Details}
We run all our experiments on a Nvidia L40S 48GB GPU card with AMD EPYC 9334 CPU and 256GB RAM.

We train all models using the AdamW optimizer with a learning rate of 1e-4 and weight decay of 1e-4. Neural baselines are trained for 100 epochs with a batch size of 64. Neurosymbolic models are trained longer, depending on the dataset: 400 epochs with batch size 64 for MultiMNIST and PokerRules, and 200 epochs with batch size 32 for CLEVR-Addition (the same batch size is also used for CoSA on CLEVR-Addition). In all cases, the best model is selected based on validation performance. All experiments are run with multiple seeds, and we report mean and standard deviation across runs in the tables.

\subsection{Datasets}

\paragraph{MultiMNIST.}
We build upon an existing MultiMNIST generator \citep{mulitdigitmnist}, modifying it to better suit the requirements of our experimental setup. In particular, we adapted the generation process to support distant supervision, where the model is trained only on the sum of the digits present in each image. The training set includes 30,000 images, and the validation and test set both include 3,000 images. Digit appearances are balanced to ensure uniform distribution across the dataset and we make sure that all possible output sums (generable from possible compositions) are seen during training. All images are normalized to the range $[-1,1]$. While the models are trained only on the total sum of the digits, we retain the individual digit annotations to support the evaluation of concept-level accuracy.

\begin{table}[H]
\centering
\caption{Example of predicates and background knowledge for MultiMNIST. This is a simplification of the version employed in the experiments, wherein a maximum of two digits are considered. It is important to note the role of $isobj\_tmp$ (objectness) in this process, as it facilitates the selection of elements to be added, thereby determining the final result, and those to be excluded.}
\scriptsize
\begin{tabular}{p{12cm}}
\toprule
$ \% \quad Neural \quad predicates $ \\


$ classifier\_slot0(X,N) :: digit\_tmp(X, 0, N) :- between(0,9,N).$ \\
$ classifier\_slot1(X,N) :: digit\_tmp(X, 1, N) :- between(0,9,N).$ \\
$ isobject\_slot0(X) :: isobj\_tmp(X, 0).$ \\
$ isobject\_slot1(X) :: isobj\_tmp(X, 1).$ \\

$ $ \\

$ \% \quad Conditioned \quad digit \quad definition $ \\
$ digit(X,ID,Y) :- \ isobj\_tmp(X, ID), \ digit\_tmp(X,ID,Y).$ \\
$ $ \\

$ \% \quad Rules $ \\
$addit(X, ID0, SumIn, SumOut) :- \ isobj\_tmp(X, ID0), digit(X, ID0, C),$ \\ $ \hspace{19em} SumOut \ is \ SumIn + C.$\\
$addit(X, ID0, ID1, SumIn, SumOut) :- \ isobj\_tmp(X, ID1), \ digit(X, ID1, C), $ \\ $ \hspace{21em} \ Y \ is \  SumIn+C, \ addit(X, ID0, Y, SumOut).$ \\

$addit(X, ID0, SumIn, SumIn) :- \ not(isobj\_tmp(X, ID0)).$\\
$addit(X, ID0, ID1, SumIn, SumOut) :- \ not(isobj\_tmp(X, ID1)),  $ \\ $ \hspace{21em} \ addit(X, ID0, SumIn, SumOut).$ \\
$ $ \\

$ addition(X, Z) :- addit(X, 0, 1, 0, Z).$ \\


\midrule
$ \% \quad Query $ \\
$ ?- \ addition(input, Z)). $ \\
\bottomrule
\end{tabular}
\label{tab:pred_multimnist}\end{table}

\paragraph{PokerRules.}
The PokerRules dataset is synthetically generated to support relational reasoning from visual input, where the label depends on the combination of card types present in the scene. The training set includes 20,000 images, with 2,000 images each for validation and testing. Each image contains between 1 and 4 playing cards (5 for the extrapolation setting) positioned along a horizontal line with slight random perturbations to their location. The cards are drawn from a fixed suit (hearts) and include the ranks: ten, jack, queen, king, and ace. Images are normalized to the range $[-1,1]$.
The training set includes five classes: nothing (1–4 cards), pair (2–4 cards), two pairs, poker and straight (4 cards). The appearance of each class is balanced across training samples. To evaluate the model's ability to generalize, we introduce three test sets. In the first, we add the unseen class \textit{three of a kind}, an out-of-distribution composition where three identical cards appear together (with 3 or 4 cards in total). In the second, we test \textit{poker-5}, where a standard poker hand is accompanied by an additional unrelated card. This introduces extrapolation to a higher number of objects (five), while keeping the underlying class structure familiar. Finally, in the third test set we evaluate \textit{full house}, a completely novel hand type composed of three cards of one rank and two of another.

\textbf{CLEVR.} Starting from the CLEVR dataset \citep{johnson2017clevr}, we construct a new dataset tailored for the CLEVR-Addition task. Images are paired randomly, and each pair is assigned as label the sum of the number of objects contained in the two images. All images are normalized to the range $[-1,1]$.
For training and validation, we use the CLEVR training split, restricting to images with at most six objects. This results in 26,132 images, with the last 10,000 reserved for validation.
All test sets are derived from the CLEVR validation split to prevent data leakage. We construct three test sets:
\begin{itemize}
    \item In-distribution: pairs of images containing at most six objects, following the same protocol as for training.
    \item Extrapolation (same class space): pairs including images with seven objects, but only those combinations whose total sum remains within the training range (i.e., $\leq 12$). This ensures neural baselines can still attempt the task.
    \item Extrapolation + OOD classes: pairs combining images with six and seven objects, producing sums beyond the training range and thus introducing new, unseen classes.
\end{itemize}

\paragraph{COCOLogic} We use the COCOLogic dataset introduced in~\cite{steinmann2025object}. To ensure consistency with the original setup, we generate the dataset using the scripts provided by the authors and verify that the resulting class distributions match those reported in the paper. COCOLogic consists of images from COCO grouped into ten classes defined by logical rules over object presence, absence, and counts (e.g., combinations or exclusions of categories such as \textit{person}, \textit{car}, or \textit{dog}). Each image satisfies exactly one rule, resulting in mutually exclusive and semantically meaningful classes. This structure makes the task inherently compositional and requires reasoning over object-level information rather than relying on global visual cues.

\begin{figure}[H]
    \centering
    \includegraphics[width=1\linewidth]{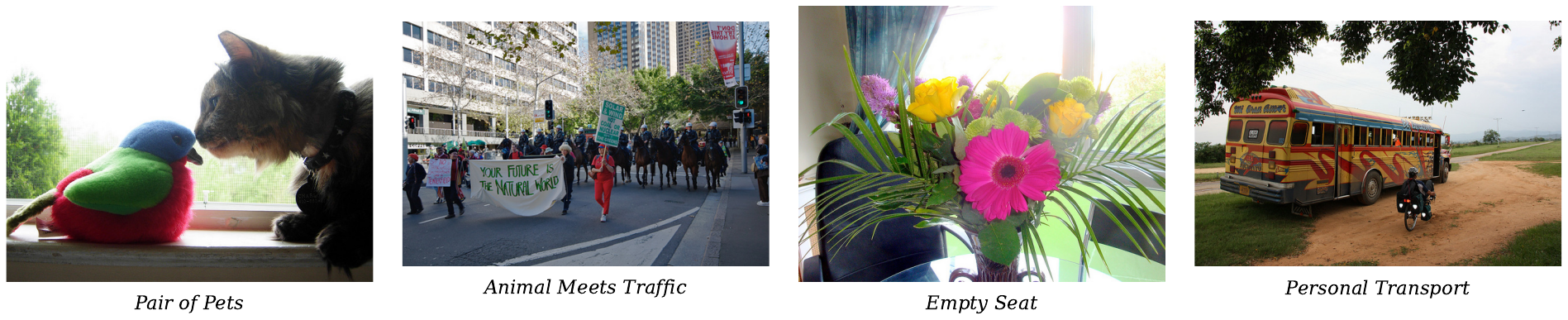}
    \caption{Example images with their corresponding classes from the COCOLogic dataset.}
    \label{fig:cocologic_samples}
\end{figure}

\section{Additional Results} \label{ap:additional}

\subsection{Complementary experiments}
This section presents complementary experiments on the MultiMNIST, PokerRules and CLEVR datasets.

\textbf{Comparison with foundation models.} To better understand how foundation models compare in our setting, we conducted an experiment using two famous Vision-Language Models: “LLaVA-1.5-7B-hf” and “Qwen2.5-VL-7B-Instruct”, on our MultiMNIST-Addition task including only out-of-distribution test set for fair comparison with our model. The results of the experiment are available in Table \ref{tab:vlm}. The VLMs were prompted with natural language queries describing the task and the possible solutions (i.e. how many digits can be in the image). Despite their general capabilities, we observe that they struggled to reliably solve the task.

\begin{table}[ht]
\centering
\caption{Accuracy results of our model and two Vision-Language Models on the MultiMNIST-addition task.}
\begin{tabular}{lccc}
\toprule
Model & OOD Compositions & Extrap. (4 digits) & Extrap. (5 digits) \\
\midrule
Llava-1.5-7b-hf & 27.43 & 2.73 & 1.87\\
Qwen2.5-VL-7B-Instruct & 59.93 & 28.67 & 14.37\\
\textbf{Ours} & \textbf{90.00}
& \textbf{69.73}
& \textbf{44.06}\\
\bottomrule
\end{tabular}
\label{tab:vlm}
\end{table}

All experiments were conducted under controlled and fixed settings (e.g., decoding temperature, and evaluation protocols) to ensure full reproducibility of results.

The prompts used for the experiments for each of the three test sets are reported below.
\begin{itemize}
    \item \textbf{OOD compositions}: \textit{"What is the value of the sum of the digits in this picture? An image with only a black background is just a 0. Reply only with the sum value. Note that the sum values can be between 0 and 27 and the image contains between 0 and 3 digits."}
    \item \textbf{Extrapolation (4 digits)}: \textit{"What is the value of the sum of the digits in this picture? An image with only a black background is just a 0. Reply only with the sum value. Note that the sum values can be between 0 and 36 and the image contains between 0 and 4 digits."}
    \item \textbf{Extrapolation (5 digits)}: \textit{"What is the value of the sum of the digits in this picture? An image with only a black background is just a 0. Reply only with the sum value. Note that the sum values can be between 0 and 45 and the image contains between 0 and 5 digits."}
\end{itemize}

\textbf{Interpolation.} In Table \ref{tab:interp}, we report results from the interpolation setting on MultiMNIST, where models are trained on images containing 0, 1, or 3 digits and tested on samples containing exactly 2 digits. As with other setups, our model demonstrates superior generalization compared to all baselines. Note that, unlike the extrapolation setting, the NeSy baseline is applicable here and provides out-of-distribution predictions that are better than the neural baselines. Additionally, the interpolation setup is particularly challenging, as the model is exposed to fewer digit configurations overall (only three instead of four, as in the standard setting), which increases the difficulty of identifying structural patterns. This suggests that providing a broader range of digit counts during training (e.g., including 4 or 5 digits) could further enhance interpolation performance.

\begin{table}[ht]
\centering
\caption{Accuracy results on the MultiMNIST-addition interpolation setting.}
\begin{tabular}{lcc}
\toprule
Model & Test (0,1,or 3 digits) & Interpolation (2 digits) \\
\midrule
CNN & 76.50${\scriptstyle \pm 0.70}$ & 24.90${\scriptstyle \pm 0.14}$ \\
SA & 98.40${\scriptstyle \pm 0.70}$ & 10.05${\scriptstyle \pm 3.74}$ \\
MESH & \textbf{99.05${\scriptstyle \pm 0.07}$} & 14.60${\scriptstyle \pm 5.09}$ \\
CoSA & 87.23${\scriptstyle \pm 5.85}$ & 22.53${\scriptstyle \pm 20.16}$ \\
NeSy & 94.35${\scriptstyle \pm 0.49}$ & 33.25${\scriptstyle \pm 7.42}$ \\
\textbf{Ours} & 89.80${\scriptstyle \pm 6.34}$ & \textbf{62.30}${\scriptstyle \pm 19.44}$ \\
\bottomrule
\end{tabular}
\label{tab:interp}
\end{table}

\textbf{Supplementary results for task generalization.} Table \ref{tab:compgen_extrap_full} presents results in an extrapolation setting involving both seen and unseen classes. The rightmost columns focus on unseen classes, where the model is tested on images containing 4 or 5 digits with sum values beyond those encountered during training. This scenario highlights the adaptability of our approach: none of the baselines can be applied here, as they lack either structural or task generalization capabilities. Our model, instead, handles both simultaneously.

\begin{table}[ht]
\centering
\caption{Accuracy results on the MultiMNIST-addition extrapolation settings. The right columns show the case where both the class and the number of digits are out of distribution.}
\begin{tabular}{lcccc}
\toprule
\multicolumn{1}{c}{} & \multicolumn{2}{c}{Extrapolation} & \multicolumn{2}{c}{OOD class extrapolation}\\
\cmidrule(lr){2-3} \cmidrule(lr){4-5}
 & 4 digits & 5 digits & 4 digits & 5 digits \\
\midrule
CNN & 22.16${\scriptstyle \pm 0.90}$ & 13.16${\scriptstyle \pm 0.96}$ & \unavailable & \unavailable \\
SA & 36.23${\scriptstyle \pm 4.44}$ & 9.76${\scriptstyle \pm 6.11}$ & \unavailable & \unavailable\\
MESH & 37.50${\scriptstyle \pm 1.15}$ & 12.00${\scriptstyle \pm 0.51}$ & \unavailable & \unavailable\\
CoSA  & 52.20${\scriptstyle \pm 14.01}$ & 25.06${\scriptstyle \pm 12.82}$ & \unavailable & \unavailable\\
\textbf{Ours} & \textbf{69.73${\scriptstyle \pm 10.74}$} & \textbf{44.06${\scriptstyle \pm 3.85}$} & \textbf{70.50${\scriptstyle \pm 9.46}$} & \textbf{46.66${\scriptstyle \pm 5.14}$}\\
\bottomrule
\end{tabular}
\label{tab:compgen_extrap_full}
\end{table}

\textbf{Concept accuracy.} In Table \ref{tab:poker_conceptaccurracy}, we provide additional metrics for the PokerRules task, including concept accuracy and card-count prediction accuracy on both the in-distribution test set and the test set with an out-of-distribution class. Unlike MultiMNIST, this task lacks an inductive bias strong enough to help the model identify individual cards using only distant supervision. 
This is a property of the program rather than of the perceptual module. The PokerRules rules only use \textit{same class} and \textit{different slots}, which are invariant under any permutation of the five ranks, so the task label determines the partition of the cards into equality classes but not which rank each class carries. Consistently, adding object-level supervision on a random 1\% of the training set breaks this symmetry and raises concept accuracy to 99.6\%.
Still, our model achieves strong accuracy in estimating the number of cards per image. We also report a best-matching concept accuracy, computed by mapping the predicted card classes to the most plausible configuration of symbols. The results reflect the increased complexity of this task, yet our model still outperforms the neurosymbolic baseline.

We also note that we tested our model's concept accuracy in the CLEVR-Addition task, achieving an excellent score of 96.28\%${\scriptstyle \pm 0.55}$. Here, the model, starting from the addition value, was able to understand the number of objects present in a single image.

\begin{table}[H]
\centering
\caption{Concept accuracy results for PokerRules}
\begin{tabular}{lcccccc}
\toprule
\multicolumn{1}{c}{} & \multicolumn{2}{c}{Task accuracy} & \multicolumn{2}{c}{Concept accuracy} & \multicolumn{2}{c}{Objects number accuracy}\\
\cmidrule(lr){2-3} \cmidrule(lr){4-5} \cmidrule(lr){6-7}
 & Test & OOD class & Test & OOD class & Test & OOD class \\
\midrule
NeSy  & 80.23${\scriptstyle \pm 2.11}$ & 0.46${\scriptstyle \pm 0.05}$ & 18.63${\scriptstyle \pm 1.37}$ & 0.13${\scriptstyle \pm 0.05}$ & 80.20${\scriptstyle \pm 5.82}$ & 76.53${\scriptstyle \pm 24.88}$\\
\textbf{Ours} & \textbf{97.90}${\scriptstyle \pm 1.17}$ & \textbf{72.23}${\scriptstyle \pm 16.72}$ & \textbf{32.00${\scriptstyle \pm 5.88}$} & \textbf{29.26${\scriptstyle \pm 13.95}$} & \textbf{81.60${\scriptstyle \pm 2.78}$} & \textbf{84.13${\scriptstyle \pm 6.60}$}\\
\bottomrule
\end{tabular}
\label{tab:poker_conceptaccurracy}
\end{table}

\begin{table}[H]
\centering
\caption{Results for CLEVR-A dataset}
\begin{tabular}{lcccccc}
\toprule
\multicolumn{1}{c}{CLEVR-A} & \multicolumn{2}{c}{Test} & \multicolumn{2}{c}{Extrapolation: 7 objects}\\
\cmidrule(lr){2-3} \cmidrule(lr){4-5}
\multicolumn{1}{c}{} & \multicolumn{2}{c}{} & \multicolumn{1}{c}{} & \multicolumn{1}{c}{OOD Class}\\
\midrule
MESH & \textbf{96.97}${\scriptstyle \pm 0.35}$ &  &  0.50${\scriptstyle \pm 0.30}$ & \unavailable \\
CoSA & 88.79${\scriptstyle \pm 3.95}$ &  &  3.44${\scriptstyle \pm 1.69}$ & \unavailable \\
\textbf{Ours} &  93.12${\scriptstyle \pm 0.56}$ &  &  \textbf{59.81}${\scriptstyle \pm 12.45}$ &  \textbf{28.57}${\scriptstyle \pm 3.71}$ \\
\bottomrule
\end{tabular}
\label{tab:clevra_results}
\end{table}

\textbf{Ablation study.} In Table \ref{tab:compgen_ablation}, we present an ablation study on the MultiMNIST-Addition task to assess the contribution of key components in our architecture. 
Each variant highlights the role of architectural choices in supporting accurate object-level reasoning and task performance. 

\begin{table}[ht]
\centering
\caption{Ablation results on the MultiMNIST-Addition task. Specifically, we evaluate performance under the following modifications: (1) replacing the object-centric encoder MESH with the standard Slot Attention mechanism; (2) using a single classifier that treats non-objectness as an additional class (common approach on object-centric learning), instead of our two-head design with separate objectness and class classifiers; (3) removing the conditioning of the class classifier on the objectness score; (4) excluding the final decoding phase and corresponding reconstruction loss during training. Regarding this last point, a recent work \citep{marconato2023not} has shown that incorporating perceptual cues through reconstruction can help align perception with symbolic reasoning, discouraging reliance on irrelevant features and preventing to fall in reasoning shortcuts.}
\begin{tabular}{lcccc}
\toprule
\multicolumn{1}{c}{} & \multicolumn{2}{c}{Task Accuracy} & \multicolumn{2}{c}{Concept Accuracy}\\
\cmidrule(lr){2-3} \cmidrule(lr){4-5}
 & Test & OOD comps & Test & OOD comps\\
\midrule
(1) Original SA & 76.91${\scriptstyle \pm 3.25}$ & 71.70${\scriptstyle \pm 3.53}$ & 30.70${\scriptstyle \pm 5.70}$ & 5.71${\scriptstyle \pm 0.14}$ \\
(2) One classifier only  & 93.49${\scriptstyle \pm 0.70}$ & 64.06${\scriptstyle \pm 19.09}$ & 15.05${\scriptstyle \pm 9.19}$ & 20.49${\scriptstyle \pm 14.14}$ \\ 
(3) No class conditioning & 92.51${\scriptstyle \pm 0.28}$ & 81.84${\scriptstyle \pm 3.46}$ & 66.77${\scriptstyle \pm 14.96}$ & 41.16${\scriptstyle \pm 12.10}$ \\
(4) No reconstruction & \textbf{95.75}${\scriptstyle \pm 1.76}$ & 89.55${\scriptstyle \pm 2.19}$ & 64.64${\scriptstyle \pm 2.33}$ & 52.20${\scriptstyle \pm 11.59}$ \\
\textbf{Ours} & 94.26${\scriptstyle \pm 2.00}$ & \textbf{90.00${\scriptstyle \pm 3.01}$} & \textbf{85.16${\scriptstyle \pm 2.45}$} & \textbf{65.46${\scriptstyle \pm 5.70}$}\\
\bottomrule
\end{tabular}
\label{tab:compgen_ablation}
\end{table}


\subsection{Training Performance}
 The processing speed in terms of iteration/sec for our model and the baselines is tabulated in Table \ref{tab:iteration_speed}. It is to be noted that speed might differ slightly with respect to the considered system and the background processes.\\
 Table \ref{tab:training_performance_tot} reports the total training time for all models in the main experiments, measured with a single seed. The project required additional computing power when testing different variations of our model and hyperparameter tuning. To mitigate the computational overhead of probabilistic reasoning, our implementation utilizes the KLay optimization framework from \cite{maene2024klay}.

\begin{table}[H]
\centering
\caption{Training speed (iterations per second, it/s) of the evaluated models across three experimental settings.}
\label{tab:iteration_speed}
\begin{tabular}{llccc}
\toprule
\textbf{Methods (↓),} & \textbf{} & \textbf{MM-A} & \textbf{PokerRules} 
& \textbf{CLEVR-A}\\
\textbf{Batch size (→)} & & \textbf{64} & \textbf{64} & \textbf{32}\\
\midrule
CNN & & 24.60 & 6.97 & N.A. \\
SA & & 14.06 & 3.72 & N.A.\\
MESH  & & 11.06 & 3.12 & 1.98\\
CoSA  & & 6.76 & 3.65 & 2.15 \\
\midrule
NeSy  & & 12.43 & 3.23 & N.A.\\
Ours & & 7.66 & 2.1 & 1.75 \\
\bottomrule
\end{tabular}
\end{table}

\begin{table}[H]
\centering
\caption{Total training time per epoch (hours:minutes) for the evaluated models across the three experimental settings.}
\begin{tabular}{llccc}
\toprule
\textbf{Methods (↓),} & \textbf{} & \textbf{MM-A} & \textbf{PokerRules}& \textbf{CLEVR-A} \\
\midrule
CNN & & $\sim$0h 31min & $\sim$1h 14min & N.A.\\
SA & & $\sim$0h 55min & $\sim$2h 20min & N.A.\\
MESH  & & $\sim$1h 10min & $\sim$2h 47min & $\sim$6h 11min\\
CoSA  & & $\sim$1h 55min & $\sim$2h 22min & $\sim$5h 03min \\
\midrule
NeSy  & & $\sim$4h 11min & $\sim$10h 46min & N.A.\\
Ours & & $\sim$6h 47min & $\sim$16h 33min & $\sim$9h 20min\\
\bottomrule
\end{tabular}
\label{tab:training_performance_tot}
\end{table}

\begin{table}[H]
\centering
\caption{Total training time of the evaluated models in the COCOLogic setting.}
\label{tab:cocologic_time}
\begin{tabular}{llcccc}
\toprule
\textbf{Methods (↓),} & \textbf{} & \textbf{} & \textbf{} 
& \textbf{}\\
\textbf{Supervision (\%) (→)} & & \textbf{0} & \textbf{30} & \textbf{60} & \textbf{100}\\
\midrule
DINOSAUR & & $\sim$5h 27min & $\sim$6h 51min & $\sim$11h 03min & $\sim$15h 40min \\
Ours & & $\sim$6h 44min & $\sim$8h 00min & $\sim$14h 33min & $\sim$19h 47min\\
\bottomrule
\end{tabular}
\end{table}


In Table \ref{tab:mma_objects_scaling}, we evaluated the scalability of our method against that of the MESH baseline in a MMA setup during training as the number of objects increases. Our model experiences a slowdown due to the complexity of the problem but we consider this as a genuine limitation, but also as the source of most of our advantages. It is exactly the marginalization over objectness and class assignments that lets a single program operate over an unknown number of objects, resolves object assignments without set matching, and provides the task-level signal that drives out-of-distribution generalization. The cost is also not intrinsic to the formulation as the symbolic layer enters the objective only through the marginal of Eq. \ref{eq:task_marginal}, so approximate inference methods can be substituted without touching perception, decoder, objective or rules. To demonstrate this, we replace exact symbolic inference with a neural approximation inspired by A-NeSI \citep{van2023nesi}. A-NeSI learns a neural prediction model that approximates the weighted model-counting operation, allowing gradients to be propagated without explicitly performing exact inference. In our setting, we exploit the finite symbolic state space to generate all possible combinations of objectness and class assignments, evaluate the program on each configuration, and use these examples to train an MLP to approximate the corresponding reasoning function. 
This experiment provides a proof of concept that the inference bottleneck can be mitigated independently of the formulation itself, with the resulting approximation proving more efficient than the baseline in some settings.

\begin{table}[H]
\centering
\caption{Number of inference iterations per second for the MM-A task (batch size 64) as the number of objects increases. We write timeout (TO) when the number of iterations per second drops below 1.}
\label{tab:mma_objects_scaling}
\begin{tabular}{llcccccccc}
\toprule
\textbf{Methods (↓),} & \textbf{} & \textbf{} & \textbf{} 
& \textbf{}\\
\textbf{Num. of objects (→)} & & \textbf{1} & \textbf{2} & \textbf{3} & \textbf{4} & \textbf{5} & \textbf{6} & \textbf{10} & \textbf{20}\\
\midrule
MESH & & 24.04 & 16.61 & 12.27 & 9.88 & 6.70 & 5.68 & 3.37 & 1.04 \\
Ours & & 16.46 & 13.24 & 7.69 & 4.06 & 2.74 & 1.70 & TO & TO\\
Ours w A-NeSI & & 20.54 & 18.00 & 14.34 & 12.67 & 10.78 & 9.21 & 4.34 & 1.19\\
\bottomrule
\end{tabular}
\end{table}

\subsection{Attention weights analysis}
We present the attention weights produced by our neurosymbolic model and by MESH on discriminative tasks trained with distant supervision. These weights are obtained from the last iteration of the slot attention module and visualized as spatial maps over the input. 
The comparison includes in-distribution, out-of-distribution, and extrapolation settings. Our model’s attention masks clearly show its ability to separate and individually attend to the objects in the scene. In contrast, the purely neural MESH model appears less capable of capturing the underlying compositional structure of the image, often focusing on the overall configuration rather than the individual objects relevant to solving the task.

Analyzing the correspondence between attention weights and the predicted objectness and classification values for each slot, we observe that slots attending to multiple regions simultaneously tend to receive low objectness scores, whereas slots with sharp, localized attention typically exhibit high objectness confidence.

To support these claims we also provide quantitative results in table \ref{tab:count_mae}. While our datasets do not naturally include ground-truth segmentation masks for standard metrics like FG-ARI, we provide a quantitative evaluation of the masks using Count Mean Absolute Error. This metric measures $| n\_active\_slots - n\_gt\_objects|$, where a slot is considered "active" if more than 1\% of its pixels exceed a threshold of $0.1$. Our model outperforms the neural baseline across all three datasets.

\begin{table}[H]
\centering
\caption{Count Mean Absolute Error for the number of active slots in the three datasets. The comparison involves our method and the baseline MESH. The lower the better.}
\begin{tabular}{lccc}
\toprule
\multicolumn{1}{c}{} & \multicolumn{1}{c}{MM-A} & \multicolumn{1}{c}{PokerRules} & \multicolumn{1}{c}{CLEVR-A}\\
\midrule
MESH  & 0.998 ${\scriptstyle \pm 0.011}$ & 1.156${\scriptstyle \pm 0.050}$ & 2.057${\scriptstyle \pm 0.037}$\\
\textbf{Ours} & \textbf{0.733}${\scriptstyle \pm 0.170}$ & \textbf{0.498}${\scriptstyle \pm 0.072}$ & \textbf{1.287}${\scriptstyle \pm 0.188}$\\
\bottomrule
\end{tabular}
\label{tab:count_mae}
\end{table}


\begin{figure}[H]
    \centering
    \includegraphics[width=0.85\linewidth]{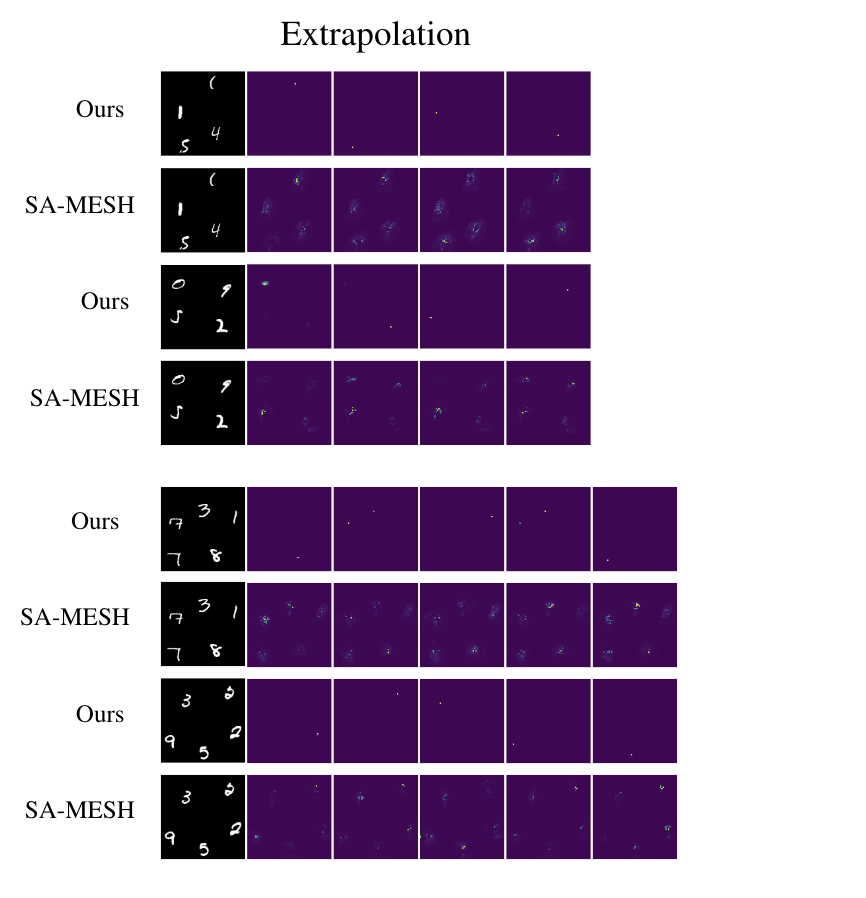}
    \caption{Attention weights of the Slot Attention mechanism for the MultiMNIST dataset.}
    \label{fig:masks2}
\end{figure}

\begin{figure}[H]
    \centering
    \includegraphics[width=0.75\linewidth]{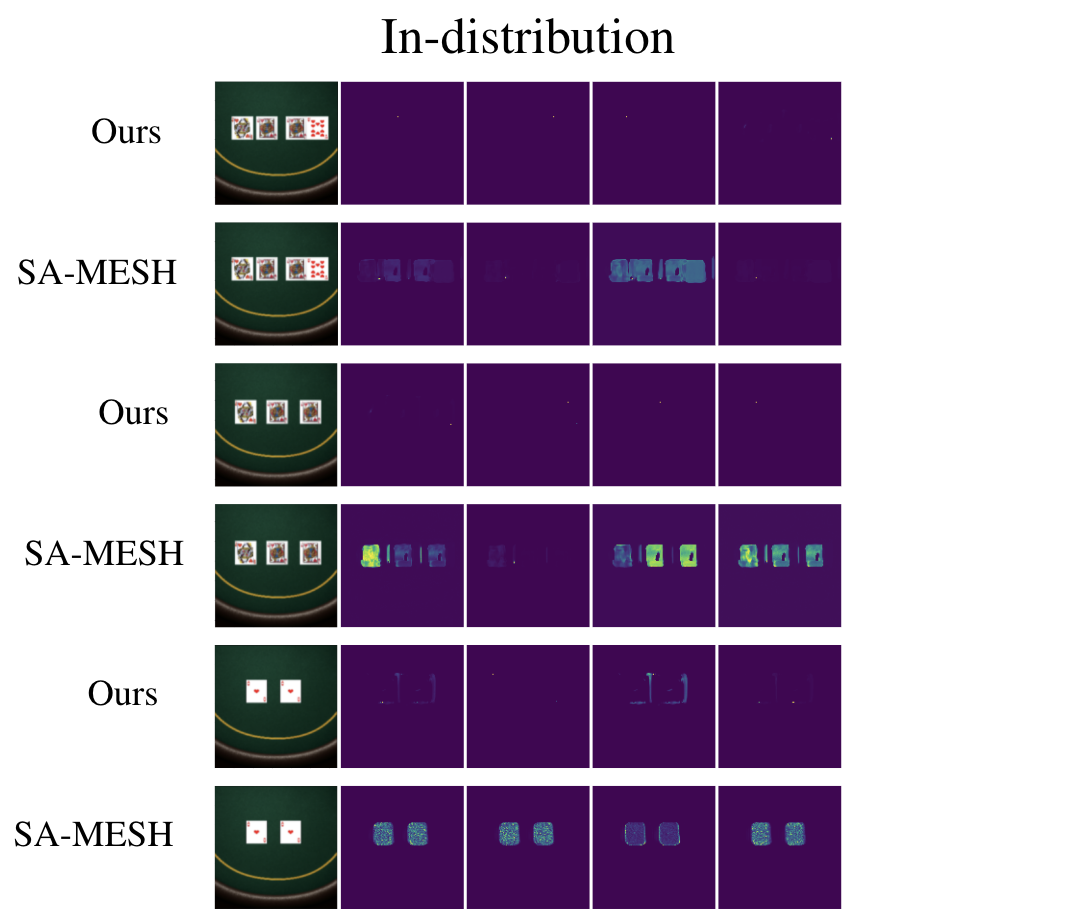}
    \includegraphics[width=0.75\linewidth]{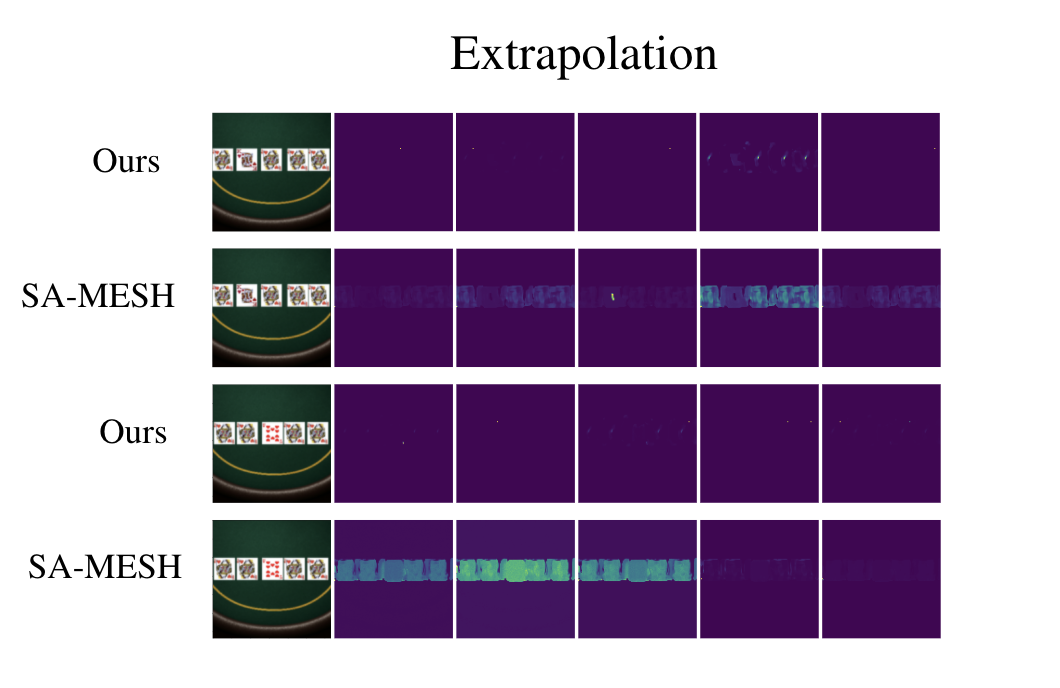}
    \caption{Attention weights of the Slot Attention mechanism for the PokerRules dataset. As the images are larger than those in MultiMNIST, it is harder to see where the attention weights are highest. However, the model is exhibiting the same behaviour as in previous experiments.}
    \label{fig:enter-label}
\end{figure}

\begin{figure}[H]
    \centering
    \includegraphics[width=0.75\linewidth]{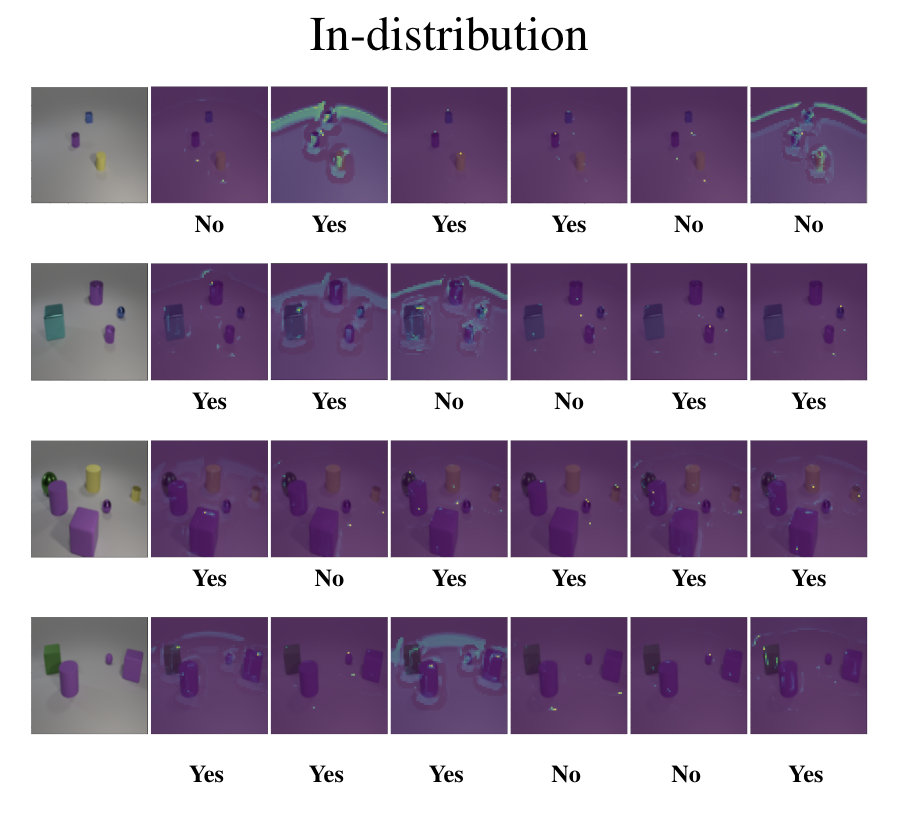}
    \includegraphics[width=0.75\linewidth]{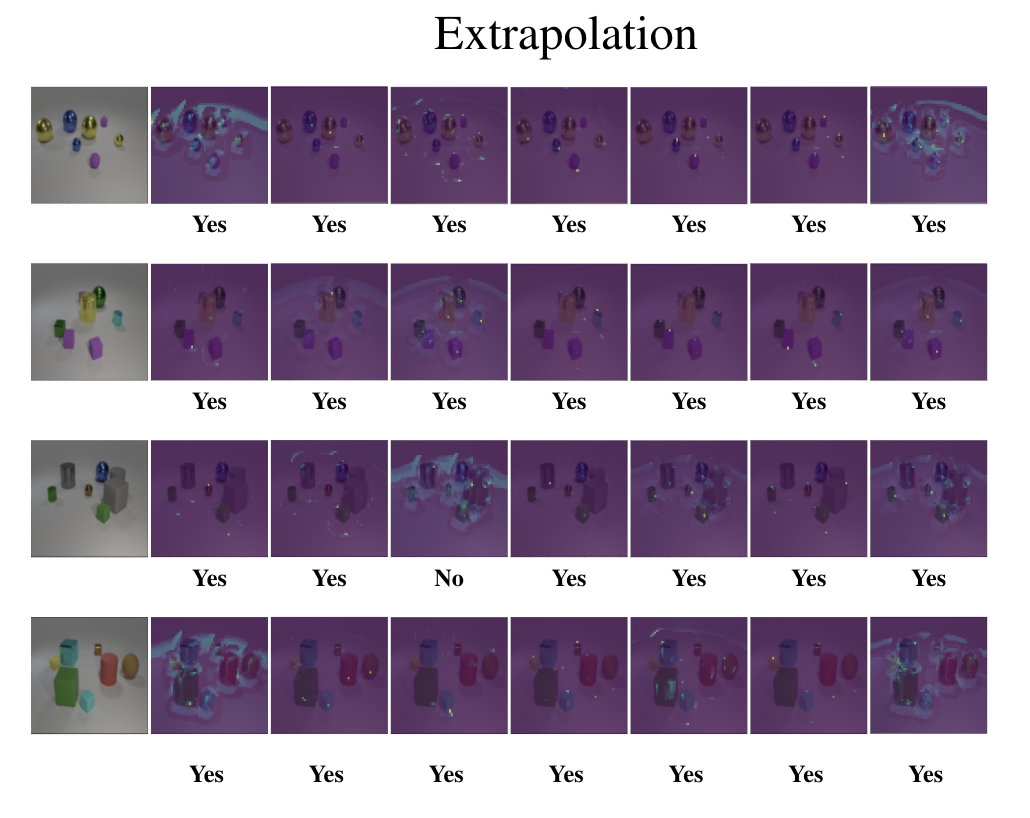}
    \caption{Attention weights of the Slot Attention mechanism trained with our method for the CLEVR dataset. As the images are larger than those in MultiMNIST, it is harder to see where the attention weights are highest. However, the model is exhibiting the same behaviour as in previous experiments. For each slot, the associated objectness score is shown, with slots above 0.5 marked as yes.}
    \label{fig:clevr_masks}
\end{figure}

\subsection{Alpha masks analysis}
In this section, we present examples of the alpha masks produced by our model during the reconstruction phase. These masks are primarily useful for reconstruction, while the attention weights are instead used to compute and refine the slot representations. Notably, the attention weights often focus on small regions within objects, whereas the alpha masks tend to approximate the full segmentation of the objects \citep{zhang2023unlocking}. 
For each alpha mask, we also report the corresponding objectness predicted by the model for that slot. The associated score is directly used in the reconstruction phase.
\begin{figure}[H]  
    \centering
    \includegraphics[width=0.60\linewidth]{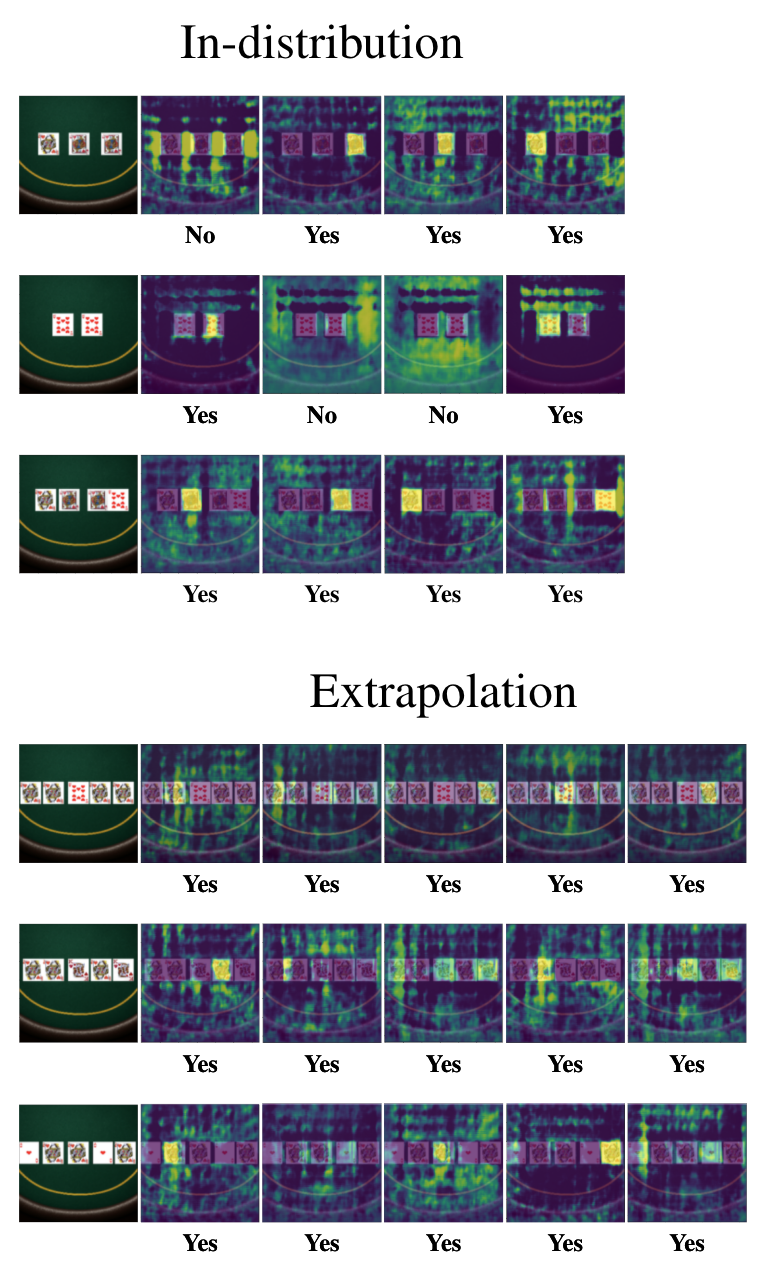}
    \caption{Alpha masks produced by the Slot Attention mechanism trained with our method on the PokerRules dataset. For each slot, the associated objectness score is shown, with slots above 0.5 marked as \textit{yes}.}
    \label{fig:placeholder}
\end{figure}



\section{Code, licenses and resources}  \label{app:licenses}

Our code is publicly available under the Apache license Version 2.0 at the following link: \href{https://github.com/StefanoColamonaco/Neurosymbolic-OCL}{https://github.com/StefanoColamonaco/Neurosymbolic-OCL}

All the data we used to build our datasets are freely available on the web with licenses:
\begin{itemize}
    \item MNIST - CC BY-SA 3.0 DEED,
    \item Cards Image Dataset - CC0: Public Domain.
    \item CLEVR - CC BY 4.0
    \item COCO - CC BY 4.0
\end{itemize}
The original datasets used as a basis for constructing our synthetic benchmarks are not redistributed. We will make the generation code of our datasets available for reproducible experiments.


\newpage
\section*{NeurIPS Paper Checklist}

The checklist is designed to encourage best practices for responsible machine learning research, addressing issues of reproducibility, transparency, research ethics, and societal impact. Do not remove the checklist: {\bf The papers not including the checklist will be desk rejected.} The checklist should follow the references and follow the (optional) supplemental material.  The checklist does NOT count towards the page
limit. 

Please read the checklist guidelines carefully for information on how to answer these questions. For each question in the checklist:
\begin{itemize}
    \item You should answer \answerYes{}, \answerNo{}, or \answerNA{}.
    \item \answerNA{} means either that the question is Not Applicable for that particular paper or the relevant information is Not Available.
    \item Please provide a short (1--2 sentence) justification right after your answer (even for \answerNA). 
\end{itemize}

{\bf The checklist answers are an integral part of your paper submission.} They are visible to the reviewers, area chairs, senior area chairs, and ethics reviewers. You will also be asked to include it (after eventual revisions) with the final version of your paper, and its final version will be published with the paper.

The reviewers of your paper will be asked to use the checklist as one of the factors in their evaluation. While \answerYes{} is generally preferable to \answerNo{}, it is perfectly acceptable to answer \answerNo{} provided a proper justification is given (e.g., error bars are not reported because it would be too computationally expensive'' or ``we were unable to find the license for the dataset we used''). In general, answering \answerNo{} or \answerNA{} is not grounds for rejection. While the questions are phrased in a binary way, we acknowledge that the true answer is often more nuanced, so please just use your best judgment and write a justification to elaborate. All supporting evidence can appear either in the main paper or the supplemental material, provided in appendix. If you answer \answerYes{} to a question, in the justification please point to the section(s) where related material for the question can be found.

IMPORTANT, please:
\begin{itemize}
    \item {\bf Delete this instruction block, but keep the section heading ``NeurIPS Paper Checklist"},
    \item  {\bf Keep the checklist subsection headings, questions/answers and guidelines below.}
    \item {\bf Do not modify the questions and only use the provided macros for your answers}.
\end{itemize}


\begin{enumerate}

\item {\bf Claims}
    \item[] Question: Do the main claims made in the abstract and introduction accurately reflect the paper's contributions and scope?
    \item[] Answer: \answerYes{} 
    \item[] Justification: The structure of our model (described in Section \ref{sec:method}), and the experiments closely follow the claimed contributions in both abstract and introduction.

\item {\bf Limitations}
    \item[] Question: Does the paper discuss the limitations of the work performed by the authors?
    \item[] Answer: \answerYes{} 
    \item[] Justification: This is done in our conclusion (Section \ref{sec:conclusions}).

\item {\bf Theory assumptions and proofs}
    \item[] Question: For each theoretical result, does the paper provide the full set of assumptions and a complete (and correct) proof?
    \item[] Answer: \answerYes{} 
    \item[] Justification: All formulae in the paper are correctly referenced and, where necessary, further explained in Appendix \ref{app:map}.

    \item {\bf Experimental result reproducibility}
    \item[] Question: Does the paper fully disclose all the information needed to reproduce the main experimental results of the paper to the extent that it affects the main claims and/or conclusions of the paper (regardless of whether the code and data are provided or not)?
    \item[] Answer: \answerYes{} 
    \item[] Justification: Details for the reproducibility of the experiments are provided in Appendix \ref{ap:expdetails}.

\item {\bf Open access to data and code}
    \item[] Question: Does the paper provide open access to the data and code, with sufficient instructions to faithfully reproduce the main experimental results, as described in supplemental material?
    \item[] Answer: \answerYes{}{} 
    \item[] Justification: A link to the code is available in Appendix~\ref{app:licenses}.

\item {\bf Experimental setting/details}
    \item[] Question: Does the paper specify all the training and test details (e.g., data splits, hyperparameters, how they were chosen, type of optimizer) necessary to understand the results?
    \item[] Answer: \answerYes{} 
    \item[] Justification: Training and test details are available in Section \ref{sec:experiments} and Appendix \ref{ap:expdetails}.

\item {\bf Experiment statistical significance}
    \item[] Question: Does the paper report error bars suitably and correctly defined or other appropriate information about the statistical significance of the experiments?
    \item[] Answer: \answerYes{} 
    \item[] Justification: We provide the standard-deviations over multiple seeds in all our tables.

\item {\bf Experiments compute resources}
    \item[] Question: For each experiment, does the paper provide sufficient information on the computer resources (type of compute workers, memory, time of execution) needed to reproduce the experiments?
    \item[] Answer: \answerYes{} 
    \item[] Justification: We reported the details of the computer resources exploited in the experiments in Appendix \ref{ap:expdetails}.
    
\item {\bf Code of ethics}
    \item[] Question: Does the research conducted in the paper conform, in every respect, with the NeurIPS Code of Ethics \url{https://neurips.cc/public/EthicsGuidelines}?
    \item[] Answer: \answerYes{} 
    \item[] Justification: Our research does not raise any ethical issue.

\item {\bf Broader impacts}
    \item[] Question: Does the paper discuss both potential positive societal impacts and negative societal impacts of the work performed?
    \item[] Answer: \answerYes{} 
    \item[] Justification: Our conclusion clearly states the broader impact of our work with reference to a more robust and interpretable AI.
    
\item {\bf Safeguards}
    \item[] Question: Does the paper describe safeguards that have been put in place for responsible release of data or models that have a high risk for misuse (e.g., pre-trained language models, image generators, or scraped datasets)?
    \item[] Answer: \answerNA{} 
    \item[] Justification: The paper poses no such risks.

\item {\bf Licenses for existing assets}
    \item[] Question: Are the creators or original owners of assets (e.g., code, data, models), used in the paper, properly credited and are the license and terms of use explicitly mentioned and properly respected?
    \item[] Answer: \answerYes{} 
    \item[] Justification: We mentioned and respected all the licences of data in Appendix \ref{app:licenses}.

\item {\bf New assets}
    \item[] Question: Are new assets introduced in the paper well documented and is the documentation provided alongside the assets?
    \item[] Answer: \answerYes{} 
    \item[] Justification: Documentation for the code and dataset generators is provided in a README file within the corresponding repository. See the answer to 'Open access to data and code'.

\item {\bf Crowdsourcing and research with human subjects}
    \item[] Question: For crowdsourcing experiments and research with human subjects, does the paper include the full text of instructions given to participants and screenshots, if applicable, as well as details about compensation (if any)? 
    \item[] Answer: \answerNA{} 
    \item[] Justification: The paper does not involve crowdsourcing nor research with human subjects.

\item {\bf Institutional review board (IRB) approvals or equivalent for research with human subjects}
    \item[] Question: Does the paper describe potential risks incurred by study participants, whether such risks were disclosed to the subjects, and whether Institutional Review Board (IRB) approvals (or an equivalent approval/review based on the requirements of your country or institution) were obtained?
    \item[] Answer: \answerNA{} 
    \item[] Justification: The paper does not involve crowdsourcing nor research with human subjects.

\item {\bf Declaration of LLM usage}
    \item[] Question: Does the paper describe the usage of LLMs if it is an important, original, or non-standard component of the core methods in this research? Note that if the LLM is used only for writing, editing, or formatting purposes and does \emph{not} impact the core methodology, scientific rigor, or originality of the research, declaration is not required.
    \item[] Answer: \answerNA{} 
    \item[] Justification: The core method development in this research does not involve LLMs as any important, original, or non-standard components. However, some VLMs are used for comparison in one of the experiments.

\end{enumerate}

\end{document}